\pdfoutput=1

\documentclass[11pt]{article}

\usepackage{times,latexsym}
\usepackage{url}
\usepackage[T1]{fontenc}
\usepackage{graphicx}
\graphicspath{ {images/} }
\usepackage{multicol}
\usepackage[dvipsnames]{xcolor}
\usepackage{xcolor}
\usepackage{times}
\usepackage{caption}
\usepackage{footmisc}
\usepackage{amsthm}
\newtheorem{theorem}{Theorem}[section]

\usepackage[labelformat=simple]{subcaption}
\usepackage{latexsym}
\usepackage{float}
\usepackage{stfloats}
\usepackage{amsmath}
\usepackage{amssymb}
\usepackage[linesnumbered,ruled,lined]{algorithm2e}
\usepackage{algpseudocode}
\usepackage{multicol}

\usepackage[final]{acl}

\usepackage{times}
\usepackage{latexsym}

\usepackage[T1]{fontenc}

\usepackage[utf8]{inputenc}

\usepackage{microtype}

\usepackage{inconsolata}

\title{Learning New Tasks from a Few Examples with Soft-Label Prototypes}

\author{
 \textbf{Avyav Kumar Singh\textsuperscript{1}},
 \textbf{Ekaterina Shutova\textsuperscript{2}},
 \textbf{Helen Yannakoudakis\textsuperscript{1}}
\\
 \textsuperscript{1}Department of Informatics, King's College London, United Kingdom 
\\
 \textsuperscript{2}ILLC, University of Amsterdam, the Netherlands
\\
 \texttt{
   avyav\_kumar.singh@kcl.ac.uk, e.shutova@uva.nl,
 }
 \\
 \texttt{helen.yannakoudakis@kcl.ac.uk}
}

\date{}

\begin{document}
\maketitle
\begin{abstract}

Existing approaches to few-shot learning in NLP rely on large language models (LLMs) and/or fine-tuning of these to generalise on out-of-distribution data. In this work, we propose a novel few-shot learning approach based on \textit{soft-label prototypes} (SLPs) designed to collectively capture the distribution of different classes across the input domain space. We focus on learning previously unseen NLP tasks from very few examples ($4$, $8$, $16$) per class and experimentally demonstrate that our approach achieves superior performance on the majority of tested tasks in this data-lean setting while being highly parameter efficient. We also show that our few-shot adaptation method can be integrated into more generalised learning settings, primarily meta-learning, to yield superior performance against strong baselines. 

\end{abstract}

\section{Introduction}

Humans have a remarkable ability to adapt knowledge gained in one domain and apply it in another setting, and to identify or disambiguate objects after observing only a handful of examples \cite{human-level-concept}. This has inspired research in few-shot learning that aims to build models that can learn a new task using only a small number of examples per class. Early few-shot learning in NLP relied on interventions at the data level, such as dataset augmentation \cite{semi-sup-training} or generation of adversarial examples from few-shot datasets \cite{miyato}, while more recent approaches \cite{van-der-heijden-etal-2021-multilingual, langedijk-etal-2022-meta} utilise meta-learning \cite{maml,proto-net} to optimise model parameters such that models adapt quickly to new tasks using past experience \cite{meta-learning-dou, wsd-holla, van-der-heijden-etal-2021-multilingual}. The advent of large language models (LLMs) has led to a plethora of further methods, including fine-tuning on different target tasks \cite{sun2020finetune, zhou-srikumar-2022-closer}, creating prompt-enhanced few-shot datasets for training \cite{prompting-gpt3,prompts-llm-cloze,parameter-prompt-tuning} as well as parameter-efficient fine-tuning methods for very large language models, with parameters running into billions \cite{lora, qlora}. 

In this paper, we propose a simple and effective approach to few-shot learning based on \textit{soft-label prototypes} (SLPs) that capture the distribution of different classes across the input domain space, inspired by previous work on generating compact representations of input training data \cite{ilia-3}. Our contributions are summarised as follows: 
1) We develop a novel neural framework for few-shot learning via soft-label prototypes that has a very small computational and memory footprint, and achieves state-of-the-art results in limited-resource settings. Our approach (DeepSLP) does not rely on (expensive) LLM parameter updates or auxiliary training data. 2) We focus on few-shot learning of new, unseen NLP tasks using as little as 4 examples per class, and demonstrate that we outperform strong baselines on the majority of test tasks. 3) We demonstrate that our approach can also be effectively adapted (MetaSLP) in high-resource settings when auxiliary training data is available for few-shot learning, and performs competitively when compared against strong baselines. 
4) We release our code and data to facilitate further research in the field.\footnote{\url{https://github.com/avyavkumar/meta-learned-lines}} 




\section{Related work}

Early few-shot learning approaches in NLP include data augmentation and semi-supervised learning; e.g., augmentation with adversarial examples \cite{miyato}, interpolation of training data into a learnable higher dimensional embedding space \cite{mixtext}, and consistency training to make models more resistant to noise \cite{consistencytraining}. 
Recent research efforts on large-scale pre-training of language models \cite{bert,gpt2,gpt3,llama,bloom,gpt4} reduce the amount of data required for their subsequent fine-tuning or utilisation in a given task. 
Instruction tuning and in-context learning \cite{gpt3,prompting-gpt3,t0, Prompting-survey,min-etal-2022-rethinking,sun-etal-2024-pearl,zhou-etal-2024-diffusion} show that natural language instructions or prompts can enhance a model's few-shot learning abilities by leveraging the language (instruction) understanding abilities of the given pretrained LLM \cite{pmlr-v139-zhao21c,liu-etal-2022-makes}. To fine-tune extremely large language models (with billions of parameters) efficiently, a host of parameter-efficient fine-tuning techniques have also been developed \cite{lora, qlora}, which leverage pre-trained LLMs and produce superior results on tasks such as question answering, reasoning, text summarisation, coding, etc.~\cite{kotitsas-etal-2024-leveraging, jiaramaneepinit-etal-2024-extreme, yang2023bayesian, ding-etal-2023-sparse}.  

However, the search space over LLMs, prompt templates and few-shot learning is so great that there is yet to be an established standard. Different models require different styles of (few-shot) prompting, and certain prompt templates may work better with specific LLMs and datasets rather than universally across the board (e.g., \citet{davis2024prompting}). Furthermore, evaluating robustness of state-of-the-art / generative LLMs on new, unseen tasks (OOD generalisation) presents a significant challenge due to their vast and unknown training data, resulting in artificially inflated performance as a result of data leakage \cite{yang2023out}.

Previous work has also tackled few-shot learning within the meta-learning paradigm of \textit{``learning to learn''} \cite{schmidhuber1987evolutionary,bengio1990learning,thrun1998learning}, utilising methods that are trained to adapt quickly (in a few gradient steps) to new tasks and from a small number of examples, using past experience. Meta-learning has emerged as a promising technique for a range of tasks \cite{maml,koch2015siamese,ravi2016optimization}, including NLP such as natural language inference, text classification, etc. \cite{obamuyide-vlachos-2019-meta,obamuyide-vlachos-2019-model,wsd-holla,smlmt-bansal,zero-shot-cross-lingual,wang-etal-2020-negative,langedijk-etal-2022-meta, mueller-etal-2023-meta}. 

In a similar spirit to parameter-efficient fine-tuning \cite{lora, qlora}, our work shifts away from the aforementioned paradigms that suffer from lack of standardisation (LLM few-shot prompting) and increased computational complexity for fine-tuning (e.g., fine-tuning extremely large language models such as GPT \cite{gpt4}. 
We present a novel, parameter-efficient few-shot learning framework (DeepSLP) based on soft-label prototypes (SLPs), which we show to be effective on a range of tasks in limited and high-resource settings, while having a substantially smaller computational and memory footprint. 
While DeepSLP does not rely on LLM fine-tuning or auxiliary training data, we present a variant (MetaSLP) that can be used for few-shot learning via auxiliary data and fine-tuned encoders. 

We target few-shot learning of new, unseen tasks (i.e., tasks and classes not previously trained on) (a) in limited-resource settings, without access to auxiliary training data, and with frozen model parameters, and (b) in high-resource settings, with access to auxiliary training data, which are used to update model parameters. For the latter, our work is similar to \citet{leopard}. The authors target few-shot learning of unseen tasks via meta-learning, utilising auxiliary training data. 



\section{Approach: few-shot learning with Soft-Label Prototypes (SLPs)} 
\label{Sec:SLP}

A soft label is defined as a vector $Y$ representing a data point's simultaneous membership to several classes \cite{ilia-1}, essentially denoting a point's partial association to different classes. Using this definition, a soft-label prototype (SLP) is defined as $(\vec{X}, Y)$, where $\vec{X}$ is a point in input space (e.g., an input feature vector) and $Y$ is its corresponding soft label. The underlying idea in \citet{ilia-1}'s work is that a small set of soft-label prototypes can be used to accurately represent a training set. 
We build on this idea and reframe SLPs for the task of few-shot learning of new, unseen tasks where very small amounts of data are available per class. 

\subsection{Generating soft-label prototypes}
Soft-label prototypes assign soft labels to every point in the input domain; therefore, a soft-label prototype at point $\vec{X}$ represents the class distribution (determined from the training data) at $\vec{X}$. The fundamental idea behind a ``soft-label'' is that, unlike hard labels, which are one-hot encoded labels, soft-labels contain a distribution of probabilistic label values at a particular point in a high-dimensional embedding space.

The process of generating soft-label prototypes from training data can be split into a two step process \cite{ilia-3}: (1) finding lines that connect the class centroids in the training data, where each line connects some of the centroids, and every centroid belongs to one line; and (2) using linear constraints to derive soft-label prototypes capturing the class distribution at the ends of each line. The two steps are presented in detail below.

\begin{figure}[t]
 \centering
 \includegraphics[width=0.4\textwidth]{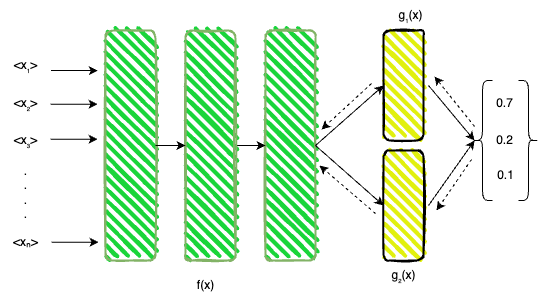}
 \caption{Learning soft-label prototypes using two trainable linear layers (yellow): example for a 3-class prototype. Dotted lines indicate backpropagation.} 
\label{fig:slp_model} %
\end{figure}

\subsubsection{Finding lines connecting all centroids}
\label{section:generate_lines}
Here we seek to find classes that lie on the same manifold. 
First, we compute the centroid of each class in the input training data. Then, we find and fit class centroids on the minimum number of lines using {recursive regression} \cite{ilia-3}. This method clusters centroids hierarchically to group similar (interval) centroids together, and fits a regression line on the centroids. The similarity of centroids within a single cluster is judged by how well all the centroids fit on a regression line. If the Euclidean distance of a particular centroid is beyond a pre-defined tolerance threshold $\epsilon$ from a line, it is removed from that cluster and assigned to another one. We use this method for all our experiments, as we experimentally find (on our dev data) that it performs well on high-dimensional data spread across many classes such as the ones we test here. 
In Appendix \ref{appendix:slp}, Figure \ref{fig:lines}, we present an example set of lines connecting all centroids. 

\subsubsection{Learning soft-label prototypes} 
Once we find the lines, we use the endpoints of each line as the location of soft-label prototypes. Therefore, for $l$ lines fitted on $n$ centroids we have $2l$ prototypes. Then, we need to find the class distribution at each end point / soft-label prototype. We develop and experiment with two different approaches to finding the class distributions, one based on constraint optimisation (\textit{constraintSLP}), and another based on gradient descent (\textit{DeepSLP}). 

\paragraph{Learning via linear constraints (constraintSLP)}
One way in which we can approach this is via constraint optimisation and, specifically, an optimisation problem that consists of two main sets of constraints \cite{ilia-3}: (i) the target class at each centroid has the maximum influence amongst all classes at certain points along the line (endpoint of the line and midpoints between classes); and (ii) the difference between the influence of the target class and the sum of the influences of all other classes along the line is maximised. 
In order to make this approach powerful enough for large, high-dimensional NLP data, we require an optimiser that scales on such complex data. To this end, we use the MOSEK solver for linear programming \cite{mosek} in the CVXPY library \cite{cvxpy} 
to perform the required computations to generate the soft-label prototype class distributions. The output of this is then a set of soft-label prototypes which ``sit'' at the ends of each line (i.e., $\vec{X}$) as shown in Appendix \ref{appendix:slp}, Figure \ref{fig:prototypes}. 

\paragraph{Learning via gradient descent (DeepSLP)}


Rather than use linear constraints to generate soft-label prototypes, we develop a novel gradient-based approach to generate soft labels as a function of an input $x$ by minimising training loss on a few-shot dataset. After generating lines connecting all class centroids (Section \ref{section:generate_lines}), we set two soft-label prototypes at the ends of the lines. Each soft-label prototype $p_i$ is denoted by $g_i(f(x))$ where $g$ is a neural network parameterised by $\theta_i$ and $f(x)$ is a point in the input space. The neural network consists of a fixed BERT \cite{bert} encoder\footnote{We use BERT as our encoder given its comparatively higher computational efficiency, and do not include LLMs such as Llama and the GPT family as they have already been pre-trained on our test tasks (found \href{https://github.com/iesl/leopard/tree/master/data/json}{here}) and hence suffer from data contamination.} given by $f(x)$, and a trainable linear layer which returns the soft-label probability distribution at any point $x$ given by $g_i(x)$. Figure \ref{fig:slp_model} presents a visual representation of our model. Compared to constraintSLP where we find soft-label probability distributions via linear constraint optimisation, we now parameterise our soft-label probability distribution with a neural network.

\begin{algorithm}[t]
\caption{DeepSLP}\label{alg:training}
$\lambda \gets$ set of lines connecting all centroids \\
$f_{\theta_l}$ is the network parameterised by $\theta_l$ for the left-end prototype on a line \\
$f_{\theta_r}$ is the network parameterised by $\theta_r$ for the right-end prototype on a line \\
$\mathcal{J} \gets$ loss function \\
$D \gets$ training data  \\
Require$\;\lambda \neq \emptyset$ \\ 
\For{$i \in \lambda$}{
\For{$epoch\;1.....N$}{
$p_{il} \gets$ location of left prototype \\
$p_{ir} \gets$ location of right prototype \\
\For{$x \in minibatch(D)$}{
$d_1 \gets$ $|| p_{il} - x ||$ \\
$d_2 \gets$ $|| p_{ir} - x ||$ \\
$pred.append \left(\frac{f_{\theta_{il}}(x)}{d_1} + \frac{f_{\theta_{ir}}(x)}{d_2}\right)$}
$d\_loss \gets \mathcal{J}(pred, D)$ \\
$loss_1 \gets \frac{d_2}{d_1+d_2} * d\_loss$ \\
$loss_2 \gets \frac{d_1}{d_1+d_2} * d\_loss$ \\
$\theta_{il} \gets \theta_{il} - \eta\nabla_{\theta_{il}}loss_1$ \\
$\theta_{ir} \gets \theta_{ir} - \eta\nabla_{\theta_{ir}}loss_2$}}
\end{algorithm}



\begin{figure}[t]
 \centering
 \includegraphics[width=0.5\textwidth]{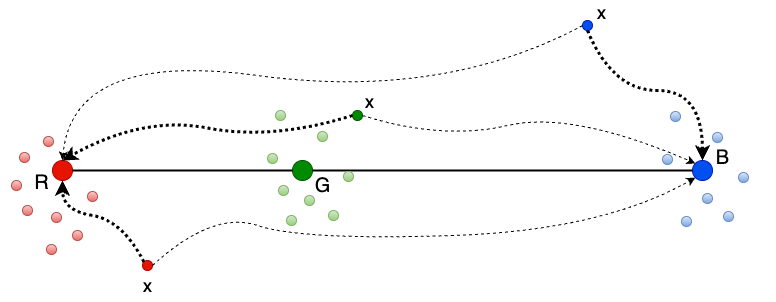}
 \caption{Training soft-label prototypes in DeepSLP. Class centroids are represented with large circles that lie on a line (\textbf{R}ed, \textbf{G}reen, \textbf{B}lue), while training set examples are represented with smaller circles of the same colour. Dotted lines represent the backpropagation error, of which the bolded ones represent a larger error per soft-label prototype. Predictions for $x$ are based on the prototypes at each end of the line.}
\label{fig:backprop}
\end{figure}

Crucially, the encoder parameters are frozen as we need our input data points to have an unchanged location in the input space -- changing their position might result in class centroids that were previously lying on a straight line to no longer lie on the line.
The model is optimised using both soft-label prototypes along a line (see Algorithm \ref{alg:training} below and Section \ref{section:class}, Equation \ref{eq:classification}). Specifically, a higher distance between a data point $x$ and a prototype leads to a correspondingly smaller effect of the prototype on the final classification; therefore, we want to penalise the prototype that is closer to $x$ more if there is an incorrect classification. Each prototype is therefore assigned a fraction of the total loss that is proportional to the other prototype's Euclidean distance from $x$. This way, the closer prototype's weights are corrected more in case of a misclassification. The complete algorithm can be seen in Algorithm \ref{alg:training}, while an example optimisation is presented in Figure \ref{fig:backprop}. 

We use cross entropy loss which gives a measure of the difference between the true and predicted labels. We initialise the weights of $g_1(x)$ and $g_2(x)$ using a uniform Xavier initialisation \cite{xavier-dist} and use warmup steps to adjust the learning rate. Epochs vary based on the number of classes in the classifier head (between 15 and 25; see datasets used in Section \ref{datasets}): preliminary experiments on the development data show that more epochs are needed when a higher number of classes lie along a line.

\subsection{Classification with soft-label prototypes}
\label{section:class} 
Given $M$ soft-label prototypes representing the input distribution of $N$ classes, we define $S = (\vec{X_1}, Y_1), ...,(\vec{X_M}, Y_M)$ to be our set of prototypes, where $\vec{X_i}$ is the location of the $i^{th}$ prototype in the input feature space and $Y_i$ is a matrix of dimension $[N\times1]$ denoting the soft labels. Given a test datapoint $\vec{x}$, we calculate the Euclidean distances ${D} = {(\vec{X_i},\vec{x})}_{i=1,2...M}$ from each prototype to $\vec{x}$. We then sort $S$ in ascending order of distances using ${D}$, weigh the probability distribution of the $i^{th}$ nearest prototype inversely by its distance to $\vec{x}$, and select the line containing the closest prototype to get $Y^*$ \cite{ilia-3}:
\begin{equation}
    Y^* = \sum_{i=1}^{k} \frac{Y_i}{d(\vec{X_i},\vec{x})} 
    \label{eq:classification}
\end{equation}
As we consider the two nearest neighbours / prototypes, we set $k$ to $2$.  $\vec{x}$ is then assigned the class $C^{SLP}(\vec{x}) = \underset{j}{\mathrm{argmax}}Y^*_j$ where $Y^*_j$ is the j\textsuperscript{th} element of $Y^*$. In other words, we sum over the k-nearest soft-label prototypes (i.e., vectors) to $\vec{x}$, and weigh each prototype inversely proportional to its distance from $\vec{x}$. $\vec{x}$ is then assigned the class with the largest value in the resulting vector (see Appendix \ref{appendix:slp} for a toy classification example).


\subsection{Meta-training DeepSLP (MetaSLP)} 

We further test the suitability of soft-labels in high-resource settings, tuning our text encoder using auxiliary training data. This is similar to the work of \cite{leopard} that develop a meta-learning approach for few-shot learning of new, unseen tasks while utilising auxiliary training data. We employ a similar approach for rapid generalisation by utilising first-order meta-learning algorithms (which we describe in detail in Appendix \ref{appendix:meta_learning}). Our model architecture is similar to DeepSLP -- it comprises a BERT encoder with two linear layers on top. We train only the last $v$ layers of our encoder to reduce computational overhead, where $v$ is a hyperparameter (See Appendix \ref{appendix:hyperparams}). We denote the encoder by $f_\theta(x)$, and each soft-label prototype at the end of a line by $g_1$ and $g_2$, parameterised by $\theta_1$ and $\theta_2$ respectively. The difference between DeepSLP and MetaSLP is that MetaSLP is trained using meta-learning (using auxiliary data described in Section \ref{datasets}), and the encoder $f_\theta(x)$ in MetaSLP is fine-tuned (as opposed to being fixed in DeepSLP), following previous work \cite{leopard}. 


\paragraph{Inner-loop training} We optimise the soft-label prototypes in the same manner as DeepSLP; i.e., we use Algorithmn \ref{alg:training} to few-shot train the linear layers $g_1(x)$ and $g_2(x)$ parameterised by $\theta_1$ and $\theta_2$ respectively. However, meta-learning requires a large set of diverse and balanced meta-learning tasks for effective learning \cite{wsd-holla}. To ameliorate this, we split the auxiliary datasets (Section \ref{datasets}) used for meta-learning into multiple pairwise tasks to meta-train MetaSLP \cite{leopard}. This means that, during training, we now consider a large number of two-class problems, as opposed to a small number of multi-class problems where the number of classes $n \geq 2$. Such a setting also enables fine-tuning of our encoder $f_\theta(x)$. In general, allowing the physical location of encodings to change (in this case via meta-learning's inner-loop training process), may result in centroids originally connected by a line to no longer be connected by that line (i.e., in the next inner-loop optimisation step). However, if we only meta-train on tasks that focus on two classes at a time, this can trivially ensure that the same line is utilised each time. 
Our inner-loop optimisation process is given in Algorithm \ref{alg:inner_loop_training} in Appendix \ref{appendix:meta_training_algo}.

\paragraph{Outer-loop training} We perform meta-learning using the updated parameters in the inner-loop training process. We experiment with both Reptile \cite{reptile} and FOMAML \cite{maml} as our meta-learning algorithms. Reptile 
can be considered a simpler variant to MAML-based meta-learning algorithms. We present our outer-loop process in Algorithm \ref{alg:outer_loop_training}, Appendix \ref{appendix:meta_training_algo}.



\paragraph{Meta-testing} We construct lines for the test sets using the trained MetaSLP model, and fine-tune them on the few-shot adaptation training data for each test task. We then use these lines for classification as described in Section \ref{section:class}.
\section{Experimental settings and datasets}
\label{datasets}

\paragraph{Experimental settings} 
We experiment with two settings in terms of amounts of available data. The first is a limited-resource setting where we only train / fine-tune our models in a few-shot manner on small amounts of training data (i.e., in the absence of auxiliary training data). The other setting is a high-resource setting where we assume that auxiliary training data is available for additional training / fine-tuning. 



\paragraph{Datasets}  
We tackle few-shot learning of previously unseen tasks (i.e., not seen during training/fine-tuning), and so our work is similar to  \citet{leopard}. 
For our high-resource setting, we train and test our models on the same data as \citet{leopard} to ensure direct comparability. For our limited-resource setting, we test in the same way but do not utilise any auxiliary training data; i.e., we only utilise few-shot fine-tuning data for unseen tasks (i.e., only using a very small set of training/fine-tuning examples for the test tasks). 


\paragraph{High-resource setting auxiliary training data} Similar to \citet{leopard}, we use GLUE \citep{glue} to train our models in the high-resource setting. This dataset consists of a range of natural language tasks such as entailment, classification and textual similarity, which are used for model training and evaluation. We use only the training split for meta-learning. Similar to \citet{leopard}, the MNLI \cite{mnli} and SNLI \cite{snli} entailment tasks, which are three-label classification problems, are split in a pairwise manner such that they are included as multiple two-label datasets during training. 
Following \citet{leopard}, we also train for detecting the sentiment contained within phrases of a sentence by using the phrase-level annotations in SST2 \cite{glue}.
We utilise the same validation sets -- labelled Amazon review data from music, toys and videos for sentiment classification \cite{amz-reviews}. We provide dataset and training details in Appendix \ref{appendix:meta_training_datasets}.

\paragraph{Evaluation data} We use the same test datasets and evaluation setting as \citet{leopard} for both the high-resource and low-resource settings. These cover a variety of text classification tasks: (a) \textit{Entity typing} -- the CoNLL-2003 \cite{conll-2003} and MIT-Restaurant \cite{mitr} datasets; (b) \textit{Review rating classification} -- review ratings from Amazon Reviews \cite{amz-reviews} with a three-way classification; (c) \textit{Text classification} -- scraped social media data from crowdflower comprising sentiment and emotion classification in a range of domains, as well as political bias detection; and (d) \textit{Natural language inference} in the scientific domain -- the SciTail dataset \cite{Khot2018SciTaiLAT}. We use the same data splits, which are publicly available. During evaluation, \citet{leopard} fine-tune their models using a small few-shot (support) training set per test task (using a $k$-shot setting of $4, 8, 16$ examples per class), and then evaluate performance on each task's dedicated, unseen test set. 
As model performance can be affected by the $k$ examples chosen for training/fine-tuning, for each task and for every k, they sample $10$ few-shot training sets and report the mean and standard deviation, which we also adopt in our experiments. 
\section{Baselines}


Our aim is to determine how well our models -- DeepSLP and MetaSLP -- perform in the low and high-resource setting respectively, compared to strong baselines when given the same few-shot adaptation sets.  
Our focus is on (a) evaluating in an extreme few-shot learning scenario where no auxiliary data is available (using DeepSLP), and (b) evaluating in a high-resource setting when additional (auxiliary) training data is available (MetaSLP). We use BERT \cite{bert} as our text encoder throughout to facilitate  model comparisons. We report DeepSLP and MetaSLP\textsubscript{REPTILE} (i.e., using Reptile as our meta-learning algorithm) in our main table of results (given their effectiveness), and present additional baselines as well as hyperparameters and training details in Appendix \ref{appendix:hyperparams}.

We use BERT \cite{bert} as our encoder as it is a text encoder that allows us to get passage-level encodings, it is computationally light-weight compared to decoder-based LLMs such as Llama \cite{llama} and GPT \cite{gpt4} (and we can carry out full fine-tuning) and, crucially, it does not suffer from data contamination as, in contrast to the more recent LLMs, it has not been pre-trained on our test data\footnote{\url{https://github.com/iesl/leopard/tree/master/data/json}}. 

\subsection{Low-resource setting baselines}

\paragraph{BERT\textsubscript{fine-tuned}} We use BERT\textsubscript{fine-tuned} reported in \citet{leopard}, which is fine-tuned (all layers) on the few-shot training set of each test task.

\paragraph{LORA\textsubscript{BERT}} LORA \cite{lora} decomposes a fine-tuned weight matrix to two low-rank matrices, which -- when multiplied and added to the original weights -- reproduce the fine-tuned weights. This is advantageous as, instead of fine-tuning all parameters, we fine-tune these two matrices with a low computational cost, as they are much smaller individually compared to fully fine-tuned weights.

\paragraph{constraintSLP} We use constraintSLP as a baseline to evaluate the effectiveness of soft-label prototypes that are based on linear constraints. 

\subsection{High-resource setting baselines}

\paragraph{Reptile} We train a meta-learning Reptile \cite{reptile} model on our auxiliary data and use it as another baseline. This allows us to directly assess the added advantage of utilising SLPs in MetaSLP\textsubscript{REPTILE}.

\paragraph{Prototypical Networks} We use ProtoNet \cite{proto-net} as another baseline \textit{for both} the high and low-resource settings. ProtoNets use Euclidean distance as a measure of similarity between points and clusters, which is similar to DeepSLP and MetaSLP that assign test points to the closest line. 

\paragraph{LEOPARD} \citet{leopard} present LEOPARD, a meta-learning algorithm that achieves the best performance across most test tasks for entity typing, ratings classification and text classification, and which we also use.


\paragraph{} We do not include HSMLMT \cite{smlmt-bansal} as a baseline as it is pretrained on semi-supervised meta-training tasks in addition to supervised learning and therefore it is not directly comparable. 

{\small \begin{table*}
\centering
\scalebox{0.75}{
\def\arraystretch{1.1}
\begin{tabular}{lllll|lll}
\hline
\textbf{Category (Classes)} &\textbf{Shot} &\textbf{LORA\textsubscript{BERT}} &\textbf{BERT\textsubscript{fine-tuned}*} &\textbf{DeepSLP} &\textbf{LEOPARD*} &\textbf{Reptile} &\textbf{MetaSLP\textsubscript{REPTILE}} \\
\hline
Political Bias (2) &4 &52.75 ± 4.33 &\colorbox{lightgray}{54.57 ± 5.02} &53.251 ± 4.042 &60.49 ± 6.66 &58.82 ± 4.31 &\colorbox{SpringGreen}{60.96 ± 6.13} \\
&8 &53.66 ± 4.25 &56.15 ± 3.75 &\colorbox{lightgray}{58.209 ± 5.198} &61.74 ± 6.73 &59.43 ± 3.79 &\colorbox{SpringGreen}{63.65 ± 4.57} \\
&16 &59.21 ± 2.27 &60.96 ± 4.25 &\colorbox{lightgray}{61.479 ± 2.974} &65.08 ± 2.14 &62.21 ± 0.72 &\colorbox{SpringGreen}{66.05 ± 1.57} \\
\hline
Emotion (13) &4 &7.56 ± 2.93 &\colorbox{lightgray}{09.20 ± 3.22} &9.076 ± 1.108 &11.71 ± 2.16 &11.65 ± 3.21 &\colorbox{SpringGreen}{11.94 ± 1.95} \\
&8 &\colorbox{lightgray}{9.02 ± 2.36} &08.21 ± 2.12 &8.041 ± 2.797 &12.90 ± 1.63 &10.56 ± 2.85 &\colorbox{SpringGreen}{13.42 ± 1.46} \\
&16 &10.29 ± 1.67 &\colorbox{lightgray}{13.43 ± 2.51} &10.919 ± 1.615 &13.38 ± 2.20 &11.62 ± 3.11 &\colorbox{SpringGreen}{14.03 ± 2.35} \\
\hline
Sentiment Books (2) &4 &51.27 ± 2.75 &54.81 ± 3.75 &\colorbox{lightgray}{58.67 ± 4.753} &82.54 ± 1.33 &76.95 ± 1.03 &\colorbox{SpringGreen}{83.22 ± 0.95} \\
&8 &58.16 ± 3.3 &53.54 ± 5.17 &\colorbox{lightgray}{64.78 ± 2.615} &83.03 ± 1.28 &77.49 ± 1.08 &\colorbox{SpringGreen}{83.8 ± 0.8} \\
&16 &59.16 ± 2.59 &65.56 ± 4.12 &\colorbox{lightgray}{67.453 ± 3.085} &83.33 ± 0.79 &77.88 ± 0.56 &\colorbox{SpringGreen}{83.8 ± 1.59} \\
\hline
Rating DVD (3) &4 &31.65 ± 4.91 &32.22 ± 08.72 &\colorbox{lightgray}{39.566 ± 5.086} &\colorbox{SpringGreen}{49.76 ± 9.80} &45.91 ± 9.85 &45.2 ± 8.91 \\
&8 &37.69 ± 3.16 &36.35 ± 12.50 &\colorbox{lightgray}{38.788 ± 4.449} &53.28 ± 4.66 &47.23 ± 9.22 &\colorbox{SpringGreen}{58.38 ± 2.9} \\
&16 &38.63 ± 5.52 &\colorbox{lightgray}{42.79 ± 10.18} &40.53 ± 4.375 &53.52 ± 4.77 &48.49 ± 8.88 &\colorbox{SpringGreen}{57.41 ± 4.71} \\
\hline
Rating Electronics (3) &4 &31.66 ± 2.94 &39.27 ± 10.15 &\colorbox{lightgray}{39.977 ± 5.959} &\colorbox{SpringGreen}{51.71 ± 7.20} &44.47 ± 8.25 &45.34 ± 7.22 \\
&8 &38.72 ± 5.95 &28.74 ± 08.22 &\colorbox{lightgray}{41.926 ± 3.985} &54.78 ± 6.48 &49.1 ± 6.81 &\colorbox{SpringGreen}{55.10 ± 5.12} \\
&16 &39.15 ± 6.6 &\colorbox{lightgray}{45.48 ± 06.13} &44.917 ± 3.164 &58.69 ± 2.41 &50.68 ± 6.8 &\colorbox{SpringGreen}{59.47 ± 2.29} \\
\hline
Rating Kitchen (3) &4 &36.63 ± 4.68 &34.76 ± 11.20 &\colorbox{lightgray}{39.624 ± 6.787} &\colorbox{SpringGreen}{50.21 ± 09.63} &45.38 ± 10.96 &45.20 ± 8.78 \\
&8 &39.69 ± 6.22 &34.49 ± 08.72 &\colorbox{lightgray}{41.081 ± 6.777} &53.72 ± 10.31 &46.71 ± 9.84 &\colorbox{SpringGreen}{54.53 ± 9.9} \\
&16 &38.17 ± 7.14 &\colorbox{lightgray}{47.94 ± 08.28} &45.801 ± 4.562 &57.00 ± 08.69 &52.87 ± 9.52 &\colorbox{SpringGreen}{58.94 ± 7.58} \\
\hline
Political Audience (2) &4 &49.75 ± 1.03 &51.02 ± 1.72 &\colorbox{lightgray}{51.741 ± 2.827} &52.60 ± 3.51 &52.45 ± 4.26 &\colorbox{SpringGreen}{54.1 ± 3.66} \\
&8 &54.05 ± 2.54 &52.80 ± 2.72 &\colorbox{lightgray}{54.506 ± 3.274} &54.31 ± 3.95 &52.87 ± 4.31 &\colorbox{SpringGreen}{56.01 ± 3.65} \\
&16 &55.39 ± 3.66 &\colorbox{lightgray}{58.45 ± 4.98} &56.956 ± 3.045 &57.71 ± 3.52 &55.6 ± 1.85 &\colorbox{SpringGreen}{58.57 ± 2.04} \\
\hline
Sentiment Kitchen (2) &4 &53.02 ± 1.54 &56.93 ± 7.10 &\colorbox{lightgray}{60.76 ± 4.426} &78.35 ± 18.36 &69.81 ± 14.58 &\colorbox{SpringGreen}{81.96 ± 3.73} \\
&8 &55.54 ± 3.47 &57.13 ± 6.60 &\colorbox{lightgray}{65.733 ± 3.198} &\colorbox{SpringGreen}{84.88 ± 1.12} &75.76 ± 1.13 &83.33 ± 1.99 \\
&16 &58.59 ± 4.83 &68.88 ± 3.39 &\colorbox{lightgray}{69.18 ± 2.589} &\colorbox{SpringGreen}{85.27 ± 1.31} &76.41 ± 0.66 &84.33 ± 1.81 \\
\hline
Disaster (2) &4 &\colorbox{lightgray}{56.02 ± 6.35} &55.73 ± 10.29 &54.252 ± 9.843 &51.45 ± 4.25 &49.76 ± 4.73 &\colorbox{SpringGreen}{55.03 ± 8.73} \\
&8 &57.46 ± 6.9 &56.31 ± 09.57 &\colorbox{lightgray}{61.3 ± 7.961} &55.96 ± 3.58 &52.17 ± 5.17 &\colorbox{SpringGreen}{57.77 ± 6.40} \\
&16 &65.79 ± 2.03 &64.52 ± 08.93 &\colorbox{lightgray}{69.28 ± 2.358} &61.32 ± 2.83 &55.37 ± 4.53 &\colorbox{SpringGreen}{65.18 ± 4.41} \\
\hline
Airline (3) &4 &24.36 ± 5.42 &42.76 ± 13.50 &\colorbox{lightgray}{50.987 ± 4.936} &54.95 ± 11.81 &57.11 ± 14.16 &\colorbox{SpringGreen}{57.39 ± 7.83} \\
&8 &52.31 ± 7.89 &38.00 ± 17.06 &\colorbox{lightgray}{55.209 ± 6.049} &61.44 ± 03.90 &64.37 ± 3.49 &\colorbox{SpringGreen}{65.67 ± 4.82} \\
&16 &54.1 ± 8.57 &58.01 ± 08.23 &\colorbox{lightgray}{60.247 ± 4.577} &62.15 ± 05.56 &66.31 ± 2.55 &\colorbox{SpringGreen}{69.48 ± 2.06} \\
\hline
Rating Books (3) &4 &34.69 ± 2.12 &39.42 ± 07.22 &\colorbox{lightgray}{42.116 ± 4.725} &54.92 ± 6.18 &\colorbox{SpringGreen}{56.57 ± 8.17} &55.79 ± 5.61 \\
&8 &39.36 ± 6.33 &39.55 ± 10.01 &\colorbox{lightgray}{42.156 ± 4.608} &59.16 ± 4.13 &57.33 ± 7.63 &\colorbox{SpringGreen}{65.74 ± 5.58} \\
&16 &41.23 ± 5.32 &43.08 ± 11.78 &\colorbox{lightgray}{46.513 ± 3.036} &61.02 ± 4.19 &63.26 ± 3.59 &\colorbox{SpringGreen}{67.87 ± 3.45} \\
\hline
Political Message (9) &4 &12.16 ± 1.46 &\colorbox{lightgray}{15.64 ± 2.73} &14.421 ± 1.095 &15.69 ± 1.57 &14.58 ± 1.78 &\colorbox{SpringGreen}{18.84 ± 1.82} \\
&8 &15.71 ± 2.04 &13.38 ± 1.74 &\colorbox{lightgray}{16.919 ± 1.756} &18.02 ± 2.32 &15.13 ± 2.16 &\colorbox{SpringGreen}{20.09 ± 2.71} \\
&16 &15.53 ± 2.55 &\colorbox{lightgray}{20.67 ± 3.89} &18.319 ± 1.74 &18.07 ± 2.41 &16.38 ± 2.15 &\colorbox{SpringGreen}{23.22 ± 1.17} \\
\hline
Sentiment DVD (2) &4 &50.77 ± 0.78 &54.98 ± 3.96 &\colorbox{lightgray}{55.003 ± 2.936} &80.32 ± 1.02 &72.03 ± 11.61 &\colorbox{SpringGreen}{80.97 ± 1.21} \\
&8 &52.24 ± 1.54 &55.63 ± 4.34 &\colorbox{lightgray}{57.527 ± 3.562} &80.85 ± 1.23 &75.79 ± 1.62 &\colorbox{SpringGreen}{81.85 ± 1.79} \\
&16 &52.6 ± 2.09 &58.69 ± 6.08 &\colorbox{lightgray}{60.76 ± 2.944} &81.25 ± 1.41 &76.69 ± 0.8 &\colorbox{SpringGreen}{83.48 ± 1.01} \\
\hline \hline
Scitail (2) &4 &43.36 ± 4.74 &\colorbox{lightgray}{58.53 ± 09.74} &54.101 ± 3.759 &\colorbox{SpringGreen}{69.50 ± 9.56} &59.13 ± 10.58 &53.48 ± 5.59 \\
&8 &54.29 ± 5.25 &\colorbox{lightgray}{57.93 ± 10.70} &56.341 ± 5.786 &\colorbox{SpringGreen}{75.00 ± 2.42} &62.63 ± 10.85 &60.79 ± 4.6 \\
&16 &52.68 ± 3.0 &\colorbox{lightgray}{65.66 ± 06.82} &59.692 ± 4.227 &\colorbox{SpringGreen}{77.03 ± 1.82} &68.03 ± 1.57 &61.67 ± 3.61 \\
\hline \hline
Restaurant (8) &4 &10.56 ± 1.36 &\colorbox{lightgray}{49.37 ± 4.28} &47.634 ± 5.237 &\colorbox{SpringGreen}{49.84 ± 3.31} &13.37 ± 2.25 &27.00 ± 2.61 \\
&8 &20.92 ± 2.4 &49.38 ± 7.76 &\colorbox{lightgray}{55.912 ± 4.494} &\colorbox{SpringGreen}{62.99 ± 3.28} &16.83 ± 3.42 &35.66 ± 2.39 \\
&16 &29.37 ± 4.05 &\colorbox{lightgray}{69.24 ± 3.68} &61.716 ± 2.208 &\colorbox{SpringGreen}{70.44 ± 2.89} &16.0 ± 3.44 &37.20 ± 2.68 \\
\hline
CoNLL (4) &4 &21.48 ± 2.71 &50.44 ± 08.57 &\colorbox{lightgray}{52.724 ± 5.84} &\colorbox{SpringGreen}{54.16 ± 6.32} &31.31 ± 5.32 &40.79 ± 3.40 \\
&8 &29.84 ± 3.28 &50.06 ± 11.30 &\colorbox{lightgray}{60.374 ± 3.731} &\colorbox{SpringGreen}{67.38 ± 4.33} &33.17 ± 5.1 &41.25 ± 5.21 \\
&16 &37.18 ± 3.32 &\colorbox{lightgray}{74.47 ± 03.10} &67.496 ± 4.551 &\colorbox{SpringGreen}{76.37 ± 3.08} &34.04 ± 3.59 &45.96 ± 4.75 \\
\hline
\end{tabular}}
\caption{
Classification performance (accuracy) of our methods (DeepSLP and MetaSLP) and baselines. * refers to the baselines as reported in \citet{leopard}. The best performing models for each setting (without and with auxiliary data) are highlighted in gray and green respectively. Double lines group similar tasks together: the first set contains intent classification tasks, the second focuses on natural language inference, and the last contains entity typing tasks.} 
\label{table:main_results}
\end{table*}}

\section{Results and Discussion}

Due to space restrictions, we  present and discuss our best models in Table \ref{table:main_results}. All other baselines and results are discussed in Appendix \ref{appendix:results}.

\paragraph{Low-resource setting} In the low-resource setting with no auxiliary data (left side of Table \ref{table:main_results}), DeepSLP outperforms all baselines, including BERT\textsubscript{fine-tuned} in $31/48$ tasks and LORA\textsubscript{BERT} in $45/48$ tasks, achieving a new state-of-the-art result. Our results demonstrate the usefulness of soft-label prototypes and their superiority over strong baselines (i.e., LLM fine-tuning and LORA / low-rank adaptation) in the low-resource setting.    

Unlike BERT\textsubscript{fine-tuned}, DeepSLP and LORA\textsubscript{BERT} do not fine-tune the encoder. Specifically, we only need to fine-tune $1500-10K$ parameters (based on the number of classes) for each line with two soft-label prototypes for DeepSLP, compared to $50K-100K$ parameters for LORA\textsubscript{BERT} with $rank=2$, and $>10^8$ parameters for BERT\textsubscript{fine-tuned}. We also note that DeepSLP is lightweight and does not require a GPU. LORA\textsubscript{BERT} mostly achieves accuracies within $90\%$ of BERT\textsubscript{fine-tuned}, in line with previous work \cite{lora, qlora} (even outperforming BERT\textsubscript{fine-tuned} in $15/48$ tasks), except for entity-typing tasks where LORA\textsubscript{BERT} struggles to generalise and achieves substantially lower performance compared to BERT\textsubscript{fine-tuned}. 

On the other hand, constraintSLP, a simpler variant of DeepSLP (see Appendix \ref{appendix:results}, Table \ref{table:appendix_deepslp} for results) is one of the lower performing baselines, together with ProtoNet. We find that constraintSLP exhibits a substantial weakness (see further details in Theorem \ref{theorem_2}, Appendix \ref{appendix:proof}): given Euclidean distance, constraintSLP does not always select the nearest class centroid to a test point. This violates our inductive bias that points located closest to a class centroid are assigned to that class. If we consider the case where $N=2$, constraintSLP essentially acts as a 1-NN with soft labels trivially at $[1, 0]$ and $[0, 1]$, with class centroids acting as the nearest neighbour. However, when generalising beyond this setting, the model's stability is affected. constraintSLP optimises soft labels using the geometric properties of a line and does not consider each (training) data point individually -- the soft labels produced by SLP are constants. DeepSLP, on the other hand, learns from training data and produces soft labels as a function of the input; therefore, it has the ability to output soft labels based on the location of an input (test) point (with the location of the prototypes being fixed). 

\paragraph{High-resource setting} MetaSLP\textsubscript{REPTILE} has the highest performance overall in text classification and entailment tasks (Tasks 1-14), with the best accuracy in $33/42$ tasks/settings. LEOPARD, on the other hand, achieves the highest score in only $8/42$. Interestingly, we find that all models in the high-resource setting have lower performance for \textit{Disaster} compared to the models in the low-resource setting. We surmise this to be due to the auxiliary data and the fact that the meta-training distribution differs substantially from the test distribution.

For entity typing tasks (CoNLL and Restaurant), LEOPARD outperforms all models, with MetaSLP\textsubscript{REPTILE} and Reptile performing comparatively poorly, even outperformed by the low-resource methods (DeepSLP and BERT\textsubscript{fine-tuned}). It should be noted that there seems to be little benefit of meta-learning with auxiliary data when tackling entity typing tasks, even for LEOPARD, as the difference between LEOPARD and  BERT\textsubscript{fine-tuned} is not substantial. 
We surmise this to be due to the fact that the meta-training distribution (i.e., GLUE tasks) is different from the test distribution for entity typing tasks which degrades performance for the test tasks. Note that we do not meta-train the entire model (only top 4 layers for Reptile and MetaSLP\textsubscript{REPTILE}), unlike LEOPARD. 

Overall, MetaSLP\textsubscript{REPTILE} outperforms all models and baselines, including Reptile in $42/48$ tasks and LEOPARD in $34/48$ tasks. Specifically, MetaSLP\textsubscript{REPTILE} consistently outperforms Reptile, demonstrating the effectiveness of our approach over its meta-learning variant (i.e., Reptile) that does not use SLPs. 
In Appendix \ref{appendix:ensemble} we present detailed analyses of DeepSLP and show that it displays several desirable properties of ensemble methods which drive its performance, in addition to it being a computationally efficient approach that only utilises a small number of parameters.  


\section{Conclusion and future work}

We presented a novel few-shot learning paradigm that is based on soft-label prototypes capturing the simultaneous membership of data points over several classes, and demonstrated its effectiveness in  low and high-resource settings. We evaluated our approach on $48$ different tasks / settings and showed that it outperforms a range of strong baselines. 
In the future, we plan to use meta-learning algorithms such as PACMAML \cite{pacmaml} and Bayesian MAML \cite{bmaml} that relax assumptions with respect to train--test set distributions and thus alleviate this current limitation in our work. 



\section{Ethics}

To the best of our knowledge, there are no ethical concerns involved in this research. We conduct our work using publicly available English datasets and tasks, and models pre-trained on English text. Our results may not generalise to other languages. To facilitate further research in the field, we release our source code and models.

\bibliography{custom}

\appendix

\section{Appendix}

\subsection{Deriving soft-label prototypes using constraintSLP}
\label{appendix:slp}

\subsubsection*{Finding lines connecting all centroids}
\begin{figure*}[hbt]
\begin{multicols}{2} 
\subfloat[Computing lines]{
    \includegraphics[width=0.5\columnwidth]{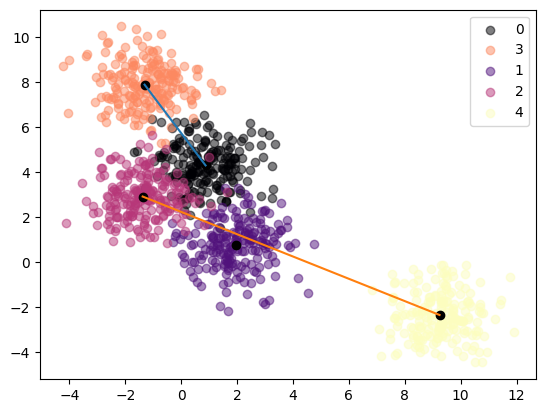}
    \label{fig:lines} 
}
\subfloat[Generating soft-label prototypes]{
    \includegraphics[width=0.5\columnwidth]{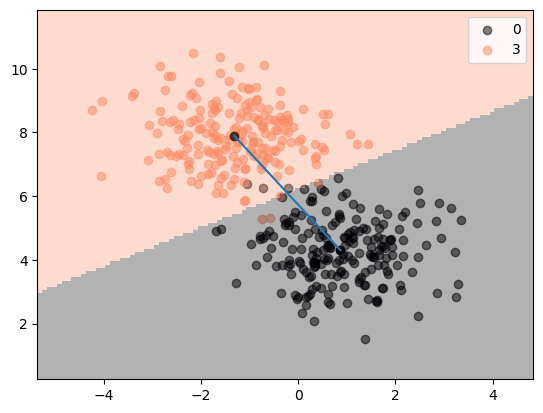}
    \includegraphics[width=0.5\columnwidth]{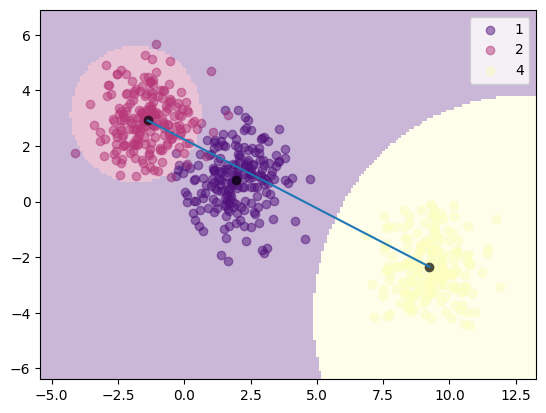}    
    \label{fig:prototypes} 
}
\subfloat[The decision landscape]{
    \includegraphics[width=0.5\columnwidth]{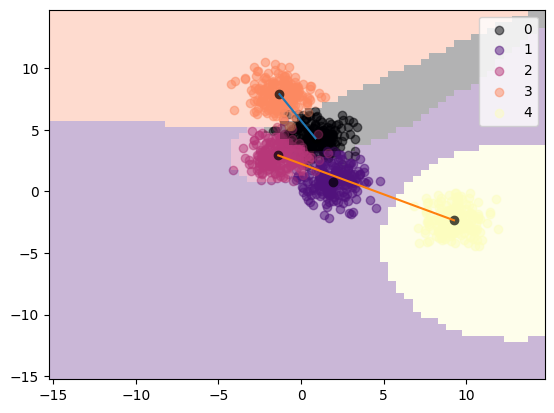}
    \label{fig:classifications}
} 
\end{multicols}
\caption{Generating and classifying data with soft-label prototypes.}
\end{figure*}

In Figure \ref{fig:lines}, we present an example set of lines connecting all class centroids. For further details on recursive regression, we refer the reader to \citet{ilia-3}.

\subsubsection*{Deriving soft-label prototypes by optimising for linear constraints}
Example soft-label prototypes which are ``set'' at the ends of each line are shown in Figure \ref{fig:prototypes}. 

\subsubsection*{Classification with constraintSLP: A toy example}

\begin{figure}[t]
    \centering
     \begin{subfigure}{0.45\columnwidth}
        \centering
        \includegraphics[scale=0.4]{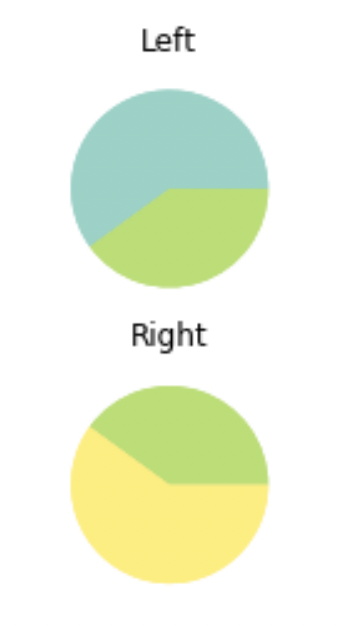}
        \caption{}
        \label{fig:diag_example_1}
     \end{subfigure}
     \hfill
     \begin{subfigure}{0.45\columnwidth}
        \centering
        \includegraphics[width=\columnwidth]{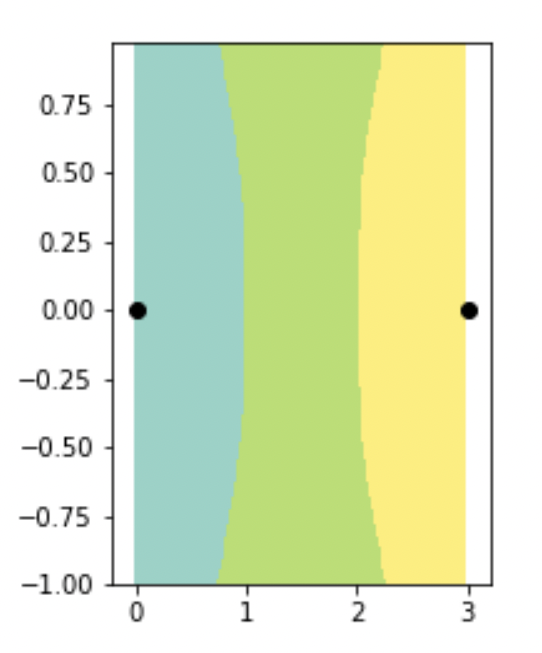}
        \caption{}
        \label{fig:diag_example_2}
     \end{subfigure}
     \hfill
\caption{Classification example with constraintSLP (figure from \citet{ilia-1}).}
\label{fig:diag_example}
\end{figure}

Figure \ref{fig:diag_example} presents an example classification with soft-label prototypes. Given the class centroids for \textit{blue}, \textit{green} and \textit{yellow} are located at $(0,0)$, $(1.5,0)$ and $(3,0)$ respectively, two soft-label prototypes are defined by a line connecting \textit{yellow} and \textit{blue}, and are thus located at $(3,0)$ and $(0,0)$ respectively. The soft labels in Figure \ref{fig:diag_example_1} contain the per-class probability distribution derived by the constraintSLP method;
for example, $p(x=blue) = 0.6$ and $p(x = green) = 0.4$ for the left prototype, and $p(x=green)= 0.4$ and $p(x=yellow) = 0.6$ for the right prototype. When a new test instance $x$ located at $(1.5, 0.8)$ is presented, we make predictions as follows: 
we find the nearest line to $x$ and consider its distance from the two prototypes at the ends of the line and multiply the class distribution of each prototype by the inverse distance as per Eq.~\ref{eq:classification}. 

Since $x$ is equidistant from both prototypes, the distance between the $x$ and each prototype is $1.5$. Therefore, the values for \textit{blue} and \textit{yellow} (given both soft-label prototypes) become $soft\_label(x = blue) = \frac{0.6}{1.5} + \frac{0}{1.5} = 0.4$ and $soft\_label(x = yellow) = \frac{0}{1.5} + \frac{0.6}{1.5} = 0.4$ respectively. In contrast, for \textit{green}, which is directly informed by both prototypes (i.e., no zero values in the numerator), the probability distribution becomes $soft\_label(x = green) = \frac{0.4}{1.5} + \frac{0.4}{1.5} = 0.53$. Therefore $x$ is classified as green. This decision boundary can be seen in Figure \ref{fig:diag_example_2}. 

\subsection{Meta-learning}
\label{appendix:meta_learning}

For encoder-based models, meta-learning has emerged as a viable methodology for few-shot learning. 
In the meta-learning paradigm, the training and test sets, referred to as $\mathcal{D}$\textsubscript{meta-train} and $\mathcal{D}$\textsubscript{meta-test}, are split into episodes. Each episode encompasses a task $\mathcal{T}$\textsubscript{i} and consists of a support set $\mathcal{D}^{(i)}$\textsubscript{support} and a query set $\mathcal{D}^{(i)}$\textsubscript{query}. Meta-learning algorithms initially fit the model on the support set of the episode (inner-loop optimisation) and then achieve generalisation across episodes by optimising performance on the query sets of the episodes (outer-loop optimisation). For evaluation, the model is first fine-tuned on the support set and then evaluated on the query set for each task $\mathcal{T}$\textsubscript{i} $\in$ $\mathcal{D}$\textsubscript{meta-test}. We describe the process algorithmically in Algorithm \ref{alg:meta-learning} and describe the \textit{MetaUpdate} process for different algorithms subsequently.

\begin{algorithm}[h]
\caption{Meta-learning}\label{alg:meta-learning}
$\alpha, \beta \leftarrow$ learning rates \\
Sample batch of tasks $\{T_i\} \sim p(T)$ \\ 
Initialise $\mathbf{\theta'_i} \leftarrow \theta$ \\
\For{$T_i \sim p(T)$}{
    Partition $T_i$ into $D^s_i$ and $D^q_i$ \\  
    $\mathbf{\theta'_i} \leftarrow \mathbf{\theta} - \alpha\nabla_\theta\mathcal{L}^s_{\mathcal{D}_i}(f_\theta)$ for $k$ steps\\ 
}
$\theta \leftarrow$ MetaUpdate$(\theta_i, D_i^q, \beta)$ \\
\end{algorithm}

\paragraph{Model Agnostic Meta-Learning} MAML \cite{maml} is an \textit{optimisation-based} meta-learning approach which incorporates generalisability across tasks in its cost function. The task loss $\mathcal{L}^q_{\mathcal{T}_i}$ is computed on the query examples in each episode, using this task-specific model. The initial model parameters $\theta$ are then updated so as to minimize the sum of the losses of all tasks in a batch, leading to improved generalisation across tasks. The \textit{MetaUpdate} step is thus defined as 

$$
\mathbf{\theta} \leftarrow \mathbf{\theta} - \beta\:\nabla_\theta\sum_{\mathcal{T}_i \sim p(\mathcal{T})}\mathcal{L}^q_{\mathcal{D}_i}(f_{\theta_i'})
$$

Note that the \textit{MetaUpdate} expression calculates the gradients of each $\theta_i$ with respect to $\theta$, thus necessitating the computation of second-order gradients. To ease computation, we use a first-order approximation of MAML (FOMAML) wherein the gradients of each $\theta_i$ are calculated with respect to $\theta_i$ and reduce the \textit{MetaUpdate} term to 

$$
\mathbf{\theta} \leftarrow \mathbf{\theta} - \beta\:\nabla_{\theta_i}\sum_{\mathcal{T}_i \sim p(\mathcal{T})}\mathcal{L}^q_{\mathcal{D}_i}(f_{\theta_i'})
$$

\paragraph{LEOPARD} 

\citet{leopard} employ meta-learning for diverse NLP tasks in an approach inspired from MAML which integrates a text encoder model with a meta-learned parameter generator to tailor task-specific initialisations for the classification head. Their inner-loop update learns the parameter generator for the task, adapts task-specific model parameters and the \textit{MetaUpdate} step adapts model parameters as done in MAML. They show that their method, LEOPARD, outperforms multi-task trained models as well as a range of other meta-learning methods. 

\paragraph{Reptile} This meta-learning algorithm, introduced by \citet{reptile}, is computationally simple compared to MAML and LEOPARD - the \textit{MetaUpdate} step simply moves the model parameters towards inner-loop fine-tuned model parameters, thus assuming the form:
$$
\theta \leftarrow \theta + \beta \frac{1}{|\{T_i\}|}\sum_{T_i \sim p(T)}(\theta_i - \theta)
$$

Despite it's simplicity, it reports strong performance on a variety of few-shot learning tasks \cite{meta-learning-dou}.

\paragraph{Prototypical Networks} Unlike optimisation-driven meta-learning methods, Prototypical Networks \cite{proto-net} is a metric-based meta-learning method that uses an embedding function $f_{\theta}$ to encode training support samples and compute a high-dimensional vector $\mu _c$ that is the arithmetic mean of the training data points of class $c$. It then uses a distance function $d$ to compute the similarity between a query instance $x$ and the mean vector of each class to get the class distribution as: 
\begin{gather*}
    p(y=c|x) = softmax(-d(f_\theta (x), \mu _c)) \\ 
    = \frac{exp(-d(f_\theta(x),\mu_c))}{\sum_{c' \in C}exp(-d(f_\theta(x),\mu_c'))}
\end{gather*}
Model optimisation is done using the loss function $J(\mathbf{\theta}) = -log(p(y = c^*|x,\theta))$.

\subsection{Analysis of soft-labels derived from linear constraints: constraintSLP}

\label{appendix:proof}
\begin{theorem} 
\label{theorem_1}
The soft-label value of each class within a single soft-label prototype generated using constraintSLP is inversely proportional to its distance from the soft-label prototype along the line connecting all classes captured by it.  
\end{theorem}
\paragraph{\textit{Proof of Theorem \ref{theorem_1} (Informal)}} Consider a line $l$ connecting three class centroids (while we focus on a three class system, the conclusions generalise to $n > 3$ classes too). The class centroids are represented by A, B and C. The soft-label prototypes at ends A and C contain the values $[a_1,a_2,a_3]$ and $[c_1,c_2,c_3]$ respectively. Consider a support example $x \in A$ at a distance $d_a$ and $d_b$ from A and C respectively. Directly using the constraints in Algorithm 4 of \citet{ilia-3}, we state that the influence of A (i.e., the distance-weighted sum of the soft-labels at $x$) should be more than the sum of the influence of the other two classes. Thus, we need to maximise: 
\begin{equation}
\frac{a_1}{d_a} + \frac{c_1}{d_b} > \left( \frac{a_2}{d_a} + \frac{c_2}{d_b} \right) + \left( \frac{a_3}{d_a} + \frac{c_3}{d_b} \right)    
\end{equation}
As we move $x$ further towards A, $d_a \rightarrow 0$ and the influence of $[a_1/d_1,a_2/d_2,a_3/d_3]$ increases thus $\sum^{i=3}_{i=2}c_i/d_b <<< \sum^{i=3}_{i=2}a_i/d_a$. Therefore we have the approximation: 
\begin{equation}
\centering
\frac{a_1}{d_a} > \frac{a_2}{d_a} +  \frac{a_3}{d_a} \implies a_1 > a_2 + a_3
\label{eq:2}
\end{equation}
If we take a support example $x \in C$, by symmetry as $x$ is moved towards C, $d_b \rightarrow 0$, we can also write: 
\begin{equation}
\centering
\frac{c_3}{d_b} >  \frac{c_2}{d_b} +  \frac{c_1}{d_b} \implies c_3 > c_2 + c_1
\label{eq:3}
\end{equation}
Furthermore, consider a point $x$ in the middle of $l$ equidistant from A and C (by a distance $d$) -- such a point will always be classified as $x \in B$. Thus, the influence of B should be higher than both A and C. Thus we have:
$$
\begin{aligned}
\frac{a_2}{d} + \frac{c_2}{d} > \frac{a_1}{d} + \frac{c_1}{d} \;\;\;\&\;\;\;
\frac{a_2}{d} + \frac{c_2}{d} > \frac{a_3}{d} + \frac{c_3}{d} \\ 
\implies a_2 + c_2 > a_1 + c_1 \;\;\;\&\;\;\; a_2 + c_2 > a_3 + c_3
\end{aligned} 
$$
From Equation \ref{eq:2} and Equation \ref{eq:3} we can replace $a_1$ and $c_3$ and get:
\begin{gather*} 
a_2 + c_2 > a_2 + a_3 + c_1 \implies c_2 > c_1   \\
a_2 + c_2 > a_3 + c_2 + c_1  \implies a_2 > a_3
\end{gather*} 
Therefore, we have:
$$
\begin{aligned}
a_1 > a_2 > a_3 \;\;\;\&\;\;\; c_3 > c_2 > c_1
\end{aligned} 
$$
This is an intuitive result as the soft-label value of each class decreases as the distance of the class centroid increases from the prototype location -- the class nearest to the prototype has the highest soft-label value and the class furthest away has the lowest soft-label value.

Recall that $\sum_{i=1}^{3}a_i = 1$ and $\sum_{i=1}^{3}c_i = 1$ and $a_i, c_i \geq 0$ $\forall i=\{1,2,3\}$ otherwise the optimisation problem becomes unbounded. Therefore, the ranges of values for $[a_1,a_2,a_3]$ and $[c_1,c_2,c_3]$ are: 
$$
\begin{aligned}
a_3 \in [0, a_2),\; a_2 \in (a_3, a_1), \; a_1 \in (a_2, 1] \\
c_3 \in (c_2, 1],\; c_2 \in (c_1, c_3), \; c_1 \in [0, c_2) 
\end{aligned} 
$$
Using Algorithm 4 in \citet{ilia-3}, we see that there are multiple constraints for $a_1$ and $a_2$ which require them to be maximised, but there are none for $a_3$. Thus, to maximise Equation \ref{eq:2}, $a_3$ adjusts to the minimum value it can get: 
$$
a_3 = min(0,a_2) = \epsilon \simeq 0
$$
By symmetry, we can also conclude that: 
$$
c_1 = min(0,c_2) = \epsilon \simeq 0
$$

These approximations are also generalisable to multiple classes connected by a line, for example, if a line connects only two centroids, the soft-labels at each end are derived as $[0,1]$ and $[1,0]$ - the same as a ``hard" label. These findings are substantiated experimentally in Table \ref{table:dist} where we examine the soft-labels generated by DeepSLP and constraintSLP using a few-shot training support set of the task \textit{airline} with $8$ examples per class – the constraintSLP soft label corresponding to the furthest class from the prototype location drops to almost zero compared to other soft label values. On the other hand, DeepSLP prevents overfitting on the nearest classes and produces a more generalised distribution of soft-labels. This is a trend generally observed in other tasks and classes as well. We further use this theorem to prove the main theorem given by Theorem \ref{theorem_2}.

\begin{figure}[t]
 \centering
 \includegraphics[width=0.5\textwidth]{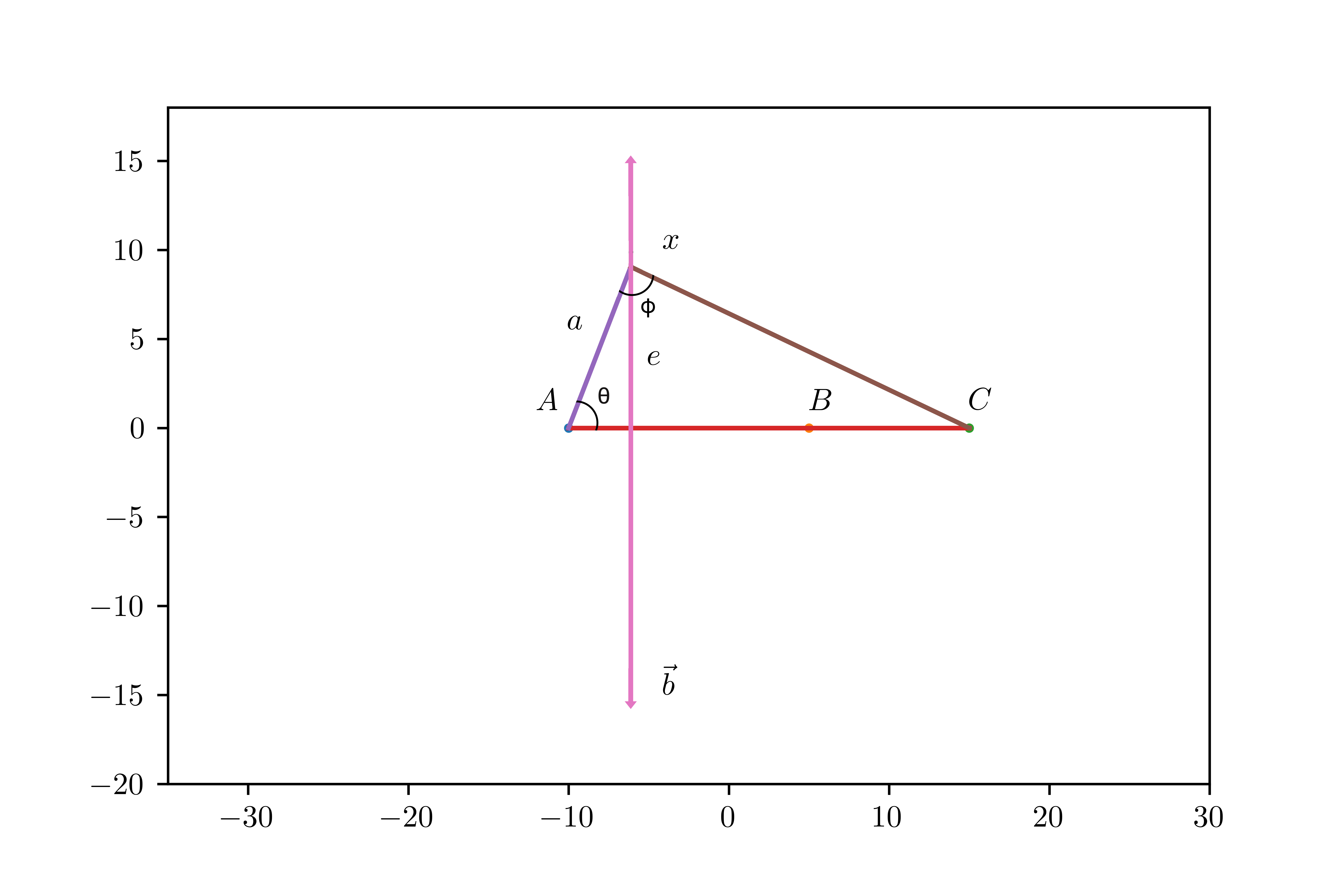}
 \caption{Schematic diagram for ascertaining $\theta$ with class centroids $A = (-10,0)$, $B=(5,0)$ and $C=(15,0)$.}
\label{fig:appendix_diag}
\end{figure}

\begin{table}[t]
\centering
\begin{tabular}{p{0.04\linewidth} p{0.38\linewidth} p{0.43\linewidth}}
\hline
\textbf{\#} & \textbf{constraintSLP} & \textbf{DeepSLP($x$)} \\ 
\hline
1 & $5.6422e-01$ & $9.7887e-01$ \\
& $4.3577e-01$ & $2.0995e-02$\\ 
& $9.7973e-16$ & $1.3500e-04$ \\
\hline
2 & $5.3090e-13$ & $2.5240e-02$ \\ 
& $4.3212e-01$ & $9.7475e-01$ \\ 
& $5.6787e-01$ & $1.0000e-05$ \\
\hline
\end{tabular}
\caption{Soft labels derived using constraintSLP and DeepSLP. \# denotes the index of the soft-label prototype lying on the line. Soft labels are constant for constraintSLP, however, they are a function of input point $x$ for DeepSLP, thus allowing more flexibility.}
\label{table:dist}
\end{table}
\begin{theorem} 
\label{theorem_2}
The constant soft-labels in constraintSLP do not always select the closest class centroid to a test point.  
\end{theorem}
\paragraph{\textit{Proof of Theorem \ref{theorem_2} (Informal)}} Furthermore, consider a line $b$ perpendicular to $l$ -- it intersects $l$ between A and B. We select $\theta$ such that $\phi = \pi/2$. We denote the complete setup diagrammatically in Figure \ref{fig:appendix_diag}. 

Consider the distance weighted influences at $x$ for class A. We have the influence as 
$\frac{a_1sin\theta}{e} + \frac{c_1cos\theta}{e} = \frac{a_1sin\theta}{e}$
as $c_1 \simeq 0$. Similarly, for class B, we have the weighted influence as  
$\frac{a_2sin\theta}{e} + \frac{c_2cos\theta}{e}$. To calculate values of $\theta$ where the weighted influence of B is more than the weighted influence of A, we get:
{\begin{gather*}
\frac{a_2sin\theta}{e} + \frac{c_2cos\theta}{e} > \frac{a_1sin\theta}{e} \\
\implies a_2sin\theta + c_2cos\theta > a_1sin\theta \\ 
\implies c_2cos\theta - (a_1 - a_2)sin\theta > 0 \\
\implies \frac{c_2cos\theta - (a_1 - a_2)sin\theta}{\sqrt{c_2^2 + (a_1 - a_2)^2}} > 0 \\
\implies cos(\theta + \alpha) > 0
\end{gather*} 

\begin{figure}[t]
    \centering
     \begin{subfigure}{\columnwidth}
         \includegraphics[width=\columnwidth]{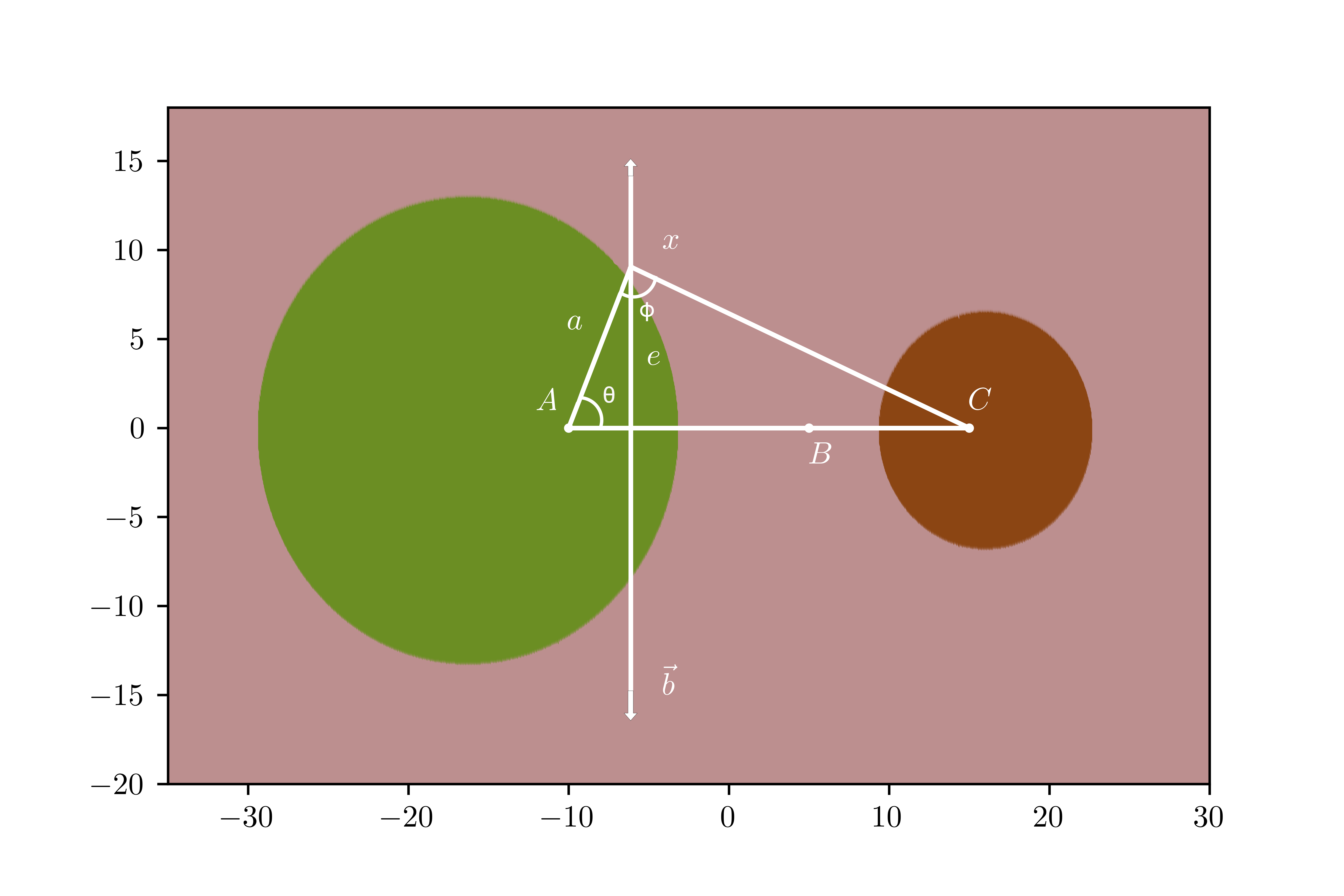}
         \caption{Denoting the decision boundaries calculated with constraintSLP. Green represents points classified as class A, pink represents the points classified 
 as class B, and brown represents the points classified as C.}
     \end{subfigure}
     \break
     \hfill
     \begin{subfigure}{\columnwidth}
         \includegraphics[width=\columnwidth]{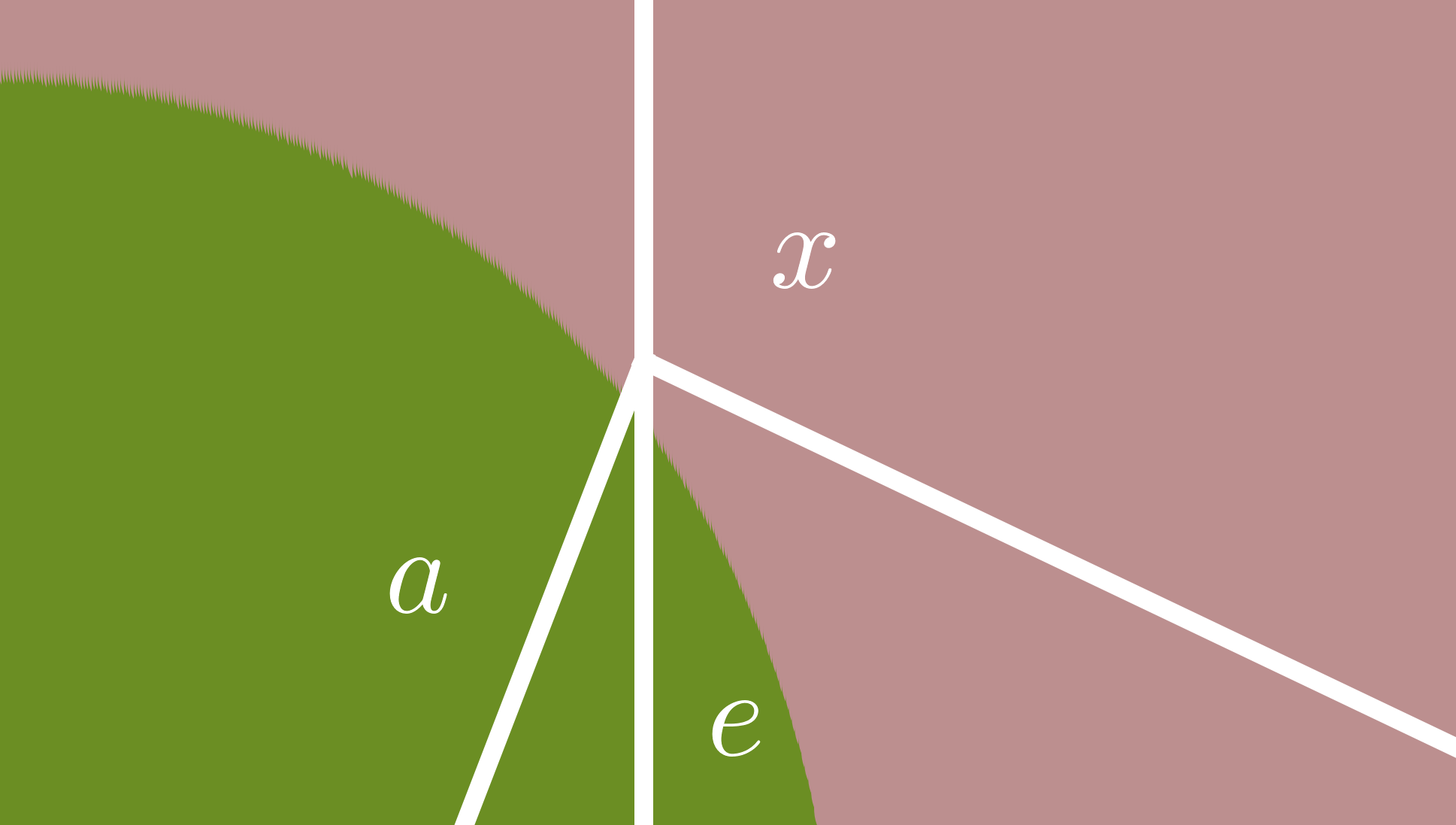}
         \caption{Zooming in at point $x$. We can see that it is classified as B.}
     \end{subfigure}
\caption{The soft-labels of the linear constraint system at A and C using constraintSLP are calculated as $[0.5963, 0.4036, 0.0001]$ and $[0.0001, 0.4495, 0.5504]$. We also get $\theta = 66.84^{\circ}$. Using these soft-labels, we calculate the decision boundaries for points in this area. We use $\theta$ to calculate the coordinates of $x$. Zooming in, we can visually inspect that $x$ is classified to class B. For $x$, the Euclidean distance of $x$ from A and B is $9.847$ and $14.338$ respectively. From the figure, we can see that constraintSLP classifies $x$ as B even though the Euclidean distance of $x$ from A is shorter.}
\label{fig:appendix_dec_boun}
\end{figure}

\noindent where $\alpha = tan^{-1}\left(\frac{a_1-a_2}{c_2}\right)$. Since $cos(\theta + \alpha) > 0$ we have $(\theta + \alpha) \in (-\pi/2,\pi/2)$ and since $\theta > 0$, thus for $\theta \in [0, \pi/2 - \alpha)$, the weighted influence of B is more than the weighted influence of A.

However, it is worth observing the result for $\theta$ derived above can contain points closer to A (using Euclidean distance) which are \textit{actually} classified as B. We can easily demonstrate this with a counter example explained in Figure \ref{fig:appendix_dec_boun}.

Therefore, for points closer to A compared to B using an Euclidean measure, constraintSLP can still return a higher value for the influence at B compared to A. This adversely affects performance in classifiers where we rely on selection of the closest class centroid for classification -- such as 1-NN, Prototypical Networks, and constraintSLP -- and we believe this is the reason behind the poor performance of constraintSLP for cases where the total classes is greater than two. 

\subsection{Training algorithms for MetaSLP}
\label{appendix:meta_training_algo}

\paragraph{Inner-loop training} Our inner-loop training algorithm for MetaSLP is presented in Algorithm \ref{alg:inner_loop_training}. The inner-loop encoder optimisation can be understood as updating the parameters of the encoder to ``push'' different classes away from each other and ``pull'' points belonging to the same class together; i.e., increase inter-class distance and decrease intra-class distance which leads to well-defined clusters per class. We present this process in Figure \ref{fig:inner-loop-opt}.

\begin{figure*}
\includegraphics[width=\textwidth]{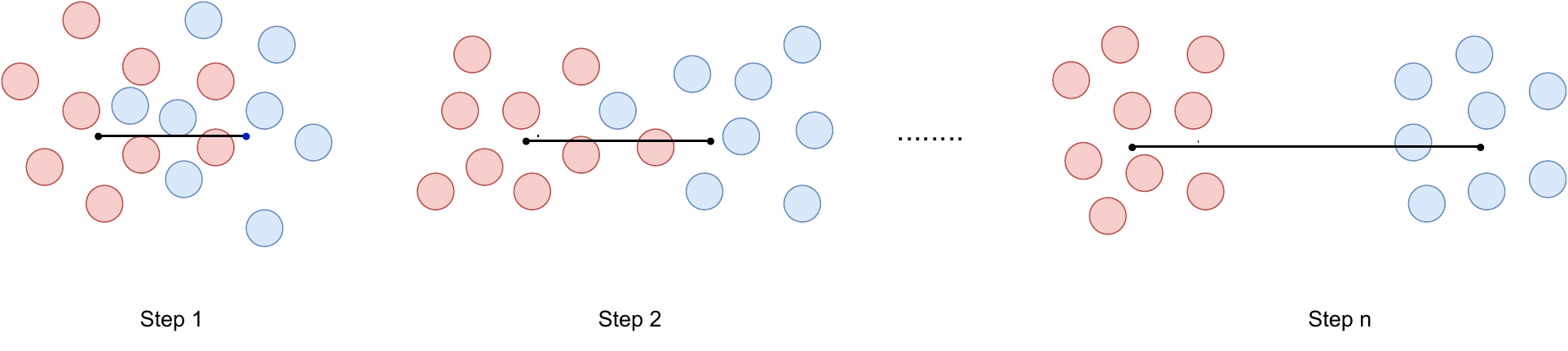}
\caption{Inner-loop training -- note that inter-class embeddings are pushed further away, and intra-class embeddings are pushed closer together across $n$ steps. The endpoints of the line mark the location of the soft-label prototypes.}
\label{fig:inner-loop-opt}
\end{figure*}

\begin{algorithm}[t]
\caption{Inner-loop training of MetaSLP}\label{alg:inner_loop_training}
$\mathcal{T}_i \gets$ meta-training task \\ 
$\mathcal{L} \gets$ line connecting both centroids of a task \\
$\alpha \gets$ inner-loop learning rate \\
$\mathcal{S} \gets$ inner-loop optimisation steps \\
Initialise $g_1x(x)$ and $g_2(x)$ randomly \\
\While{$s < \mathcal{S}$}{
Sample support examples $X^s$ for $\mathcal{T}_i$ \\ 
Calculate locations of each soft-label prototype in $\mathcal{L}$\\ 
Use Equation \ref{eq:classification} to classify $x \in X^s$ \\
Calculate $\nabla\mathcal{L}_{\mathcal{T}_i}(f_\theta(x), g_1(x), g_2(x))$ \\
Scale $\nabla_\phi\mathcal{L}_{\mathcal{T}_i}(g_1(x))$ and $\nabla_\omega\mathcal{L}_{\mathcal{T}_i}(g_2(x))$ by the distances from the soft-label prototypes \\ 
$\theta'_i \gets \theta - \alpha\nabla_\theta\mathcal{L}_{\mathcal{T}_i}(f_\theta(x))$ \\ 
$\theta'_{1_i} \gets \theta_1 - \alpha\nabla_\phi\mathcal{L}_{\mathcal{T}_i}(g_1(x))$ \\ 
$\theta'_{2_i} \gets \theta_2 - \alpha\nabla_\omega\mathcal{L}_{\mathcal{T}_i}(g_2(x))$ 
} 
\end{algorithm}

\paragraph{Outer-loop training} Our outer-loop training algorithm is presented in Algorithm \ref{alg:outer_loop_training}.

\begin{algorithm}[t]
\caption{Outer-loop training of MetaSLP}\label{alg:outer_loop_training}
$\mathcal{T} \gets$ batch of meta-training tasks, $|\mathcal{T}| = n$ \\
$\mathcal{M} \gets$ batch of distinct inner-loop optimised models parameterised by $\theta_i$, $|\mathcal{M}| = n$ \\
$\beta \gets$ outer-loop learning rate \\
\For{$\mathcal{T}_i,\; \mathcal{M}_i \in \mathcal{T},\; \mathcal{M}$}{
\If{FOMAML}{
Sample query examples $X^q$ for $\mathcal{T}_i$, $X^s \cap X^q = \Phi$ \\
Use Equation \ref{eq:classification} to classify $x \in X^q$ \\
Calculate $\nabla_{\theta'_i}\mathcal{L}_{\mathcal{T}_i}(f_{\theta_i'}(x), g_{\phi_i'}(x), h_{\omega_i'}(x))$ \\
Update $ \nabla_\theta\mathcal{L}_{\mathcal{T}_i}(f_\theta(x)) += \nabla_{\theta'_i}\mathcal{L}_{\mathcal{T}_i}(f_{\theta_i'}(x), g_{\phi_i'}(x), h_{\omega_i'}(x))$
}
\If{Reptile}{
Update $ \nabla_\theta\mathcal{L}_{\mathcal{T}_i}(f_\theta(x)) += \theta - \theta_i$
}
} 
Update $\theta \gets \theta - \frac{\beta}{n}\sum^n_{i=1}\nabla_\theta\mathcal{L}_{\mathcal{T}_i}(f_\theta(x))$
\end{algorithm}

\subsection{Meta-training details}
\label{appendix:meta_training_datasets}

GLUE \cite{glue} tasks and their details are provided in Table \ref{table:glue}. These tasks include MNLI \cite{mnli}, SST2 \cite{sst2}, CoLA \cite{cola}, MRPC \cite{mrpc}, QQP \cite{qqp}, QNLI \cite{glue}, RTE \cite{rte} and SNLI \cite{snli}. We employ the same tasks as \citet{leopard} to ensure direct comparability. Note that the datasets and classes in GLUE are completely different from the datasets used for evaluating the model - thus the final model fine-tunes on unseen few-shot data and learns classes it has previously not encountered. 

To train our model to detect the sentiment contained within phrases of a sentence by using the annotations for phrases within sentences for SST2, we append a separator token and the annotated phrase for each sentence at the end of the sentence in the form ``[CLS] <sentence\_1> [SEP] <sentence\_2> [SEP]" and obtain the passage level embedding for training. 

\begin{table}[t]
\centering
\scalebox{0.85}{
\def\arraystretch{1.1}
\begin{tabular}{p{0.2\linewidth}p{0.1\linewidth}p{0.2\linewidth}p{0.2\linewidth}p{0.1\linewidth}}
\hline
\textbf{Dataset} &\textbf{Labels} &\textbf{Training Size} &\textbf{Validation Size} &\textbf{Test Size} \\
\hline
CoLA & 2 & 8551 & 1042 & - \\
MRPC & 2 & 3669 & 409 & - \\
QNLI & 2 & 104744 & 5464 & - \\
QQP & 2 & 363847 & 40431 & - \\
RTE & 2 & 2491 & 278 & - \\
SNLI & 3 & 549368 & 9843 & - \\
SST-2 & 2 & 67350 & 873 & - \\
MNLI & 3 & 392703 & 19649 & - \\
\hline
\end{tabular}}
\captionof{table}
[Details of GLUE datasets used for meta-training the soft-label model]{Details of GLUE tasks used for meta-training.}
\label{table:glue}
\end{table}

\subsection{Hyperparameters}
\label{appendix:hyperparams}

\paragraph{Generating lines}
\label{section:hyperparameters_linear}
The hyperparameters used to generate lines are: (a) $\epsilon$, which is a control factor used to denote the maximum tolerance between a centroid and the line assigned to it using Euclidean distance---we use a tolerance value of $1e-1$; and (b) $l$, which denotes the maximum number of lines used to connect all centroids. We experiment with a range of values ($l \in \{0.25n, 0.5n, 0.75n, n-1\}$, where $n$ is the number of centroids), but find  $l=\lceil n/2 \rceil$ to give the best accuracy on the validation data with the minimum number of lines required.\footnote{The right choice of hyperparameters is key as the optimisation process fails when it is not possible to connect $n$ centroids with $l$ lines. } 

\paragraph{DeepSLP}
\label{section:hyperparameters_soft_labels}
For DeepSLP, we find that more epochs are needed to train models with a higher number of soft labels (i.e., a higher number of classes in the output of the classifier head) - essentially, 3 classes fitted on a line need more epochs compared to 2 classes fitted on a line. We use \textit{AdamW} \cite{adamw} as our optimiser and perform hyperparameter tuning on the validation set. We only need a few epochs (5 to 10) to generalise well depending on the training task. We fix a random seed, train our models and evaluate performance on the test tasks. We repeat this process across three different seeds and report the mean and standard deviation. Hyperparameters for all baselines in this setting can be found online in our code repository\footnote{\url{https://github.com/avyavkumar/meta-learned-lines}}.

\paragraph{Meta-training} For inner-loop optimisation, we use SGD as an optimiser with Nesterov and a momentum factor. We use a cosine annealing learning rate scheduler \citep{cosine_anneal} on our outer-loop learning rate to decay the learning rate from a starting rate to an end rate without restarts across one epoch for MetaSLP\textsubscript{REPTILE}. We use AdamW \citep{adamw} as our outer-loop optimiser with AMSGrad \citep{amsgrad}. We employ early stopping and stop training if our model does not improve it's validation set accuracy over 100 batches. We use learnable inner-loop learning rates for MetaSLP\textsubscript{FOMAML} per parameter group for better optimisation as indicated by previous literature \cite{howtotrainmaml}. All meta-training hyperparameters can be found in Appendix \ref{appendix:hyperparams}, Table \ref{table:hyperparam}.

\paragraph{Meta-testing} Similar to inner-loop optimisation at meta-training, we use SGD with Nesterov and the same optimiser hyperparameters. However, we decay the learning rate using a cosine scheduler across all fine-tuning epochs to prevent overfitting on the few-shot (support) training set per task for MetaSLP\textsubscript{REPTILE}.

\begin{table*}[t]
\centering
\scalebox{0.85}{
\def\arraystretch{1.1}
\begin{tabular}{p{0.2\linewidth}p{0.3\linewidth}p{0.17\linewidth}p{0.17\linewidth}p{0.13\linewidth}}
\hline
\textbf{Parameter} &\textbf{Search Space} &\textbf{MetaSLP\textsubscript{REPTILE}} &\textbf{MetaSLP\textsubscript{FOMAML}} &\textbf{Reptile}\\
\hline
Tunable layers ($v$) & $[1,2,3,4]$ & $4$ & $4$ & $4$ \\
K-shot & $[8,16,32]$ & $16$ & $16$ & $16$\\
Batch size & $[8,16,32]$ & $16$ & $16$ & $16$\\
Steps & $[3,5,7]$ & $5$ & $5$ & $5$\\
$\alpha_f$ & $[5e-3,1e-3,1e-4]$ & $5e-3$ & Learnable & $1e-3$ \\
$\alpha_g,\;\alpha_h$ & $[5e-3,1e-3,1e-4,1e-2]$ & $5e-3$ & Learnable & $1e-3$ \\
Nesterov & $[True, False]$ & $True$ & $True$ & $True$\\
Momentum & $[0.5,0.7,0.9]$ & $0.9$ & $0.9$ & $True$\\
$\beta$\textsubscript{initial} & $[1e-5,2e-5,5e-5]$ & $5e-5$ & $2e-5$ & $1e-5$ \\
$\beta$\textsubscript{final} & $[1e-5,2e-5,5e-6]$ & $2e-5$ & $2e-5$ & $1e-5$ \\
Task sampling & [\textit{square root, uniform}] & \textit{square root} & \textit{square root} & \textit{square root} \\
\hline
\end{tabular}}
\captionof{table}
[Meta-learning hyperparameters]{Meta-training hyperparameters.}
\label{table:hyperparam}
\end{table*}

\subsection{Results} 
\label{appendix:results}

The complete set of results of all models and baselines can be seen in Table \ref{table:appendix_deepslp} for the low-resource setting and DeepSLP, and Table \ref{table:appendix_metaslp} for the high-resource setting and MetaSLP. 

\paragraph{DeepSLP} Our results (Table \ref{table:appendix_deepslp}) demonstrate that DeepSLP\textsubscript{BERT} outperforms BERT\textsubscript{fine-tuned} in $31/48$ tasks, constraintSLP\textsubscript{BERT} in $43/48$ tasks and LORA\textsubscript{BERT} in $45/48$ tasks, demonstrating the usefulness of soft-label prototypes and superiority over the ``standard'' LLM fine-tuning paradigm, as well as the simpler constraintSLP variant. constraintSLP\textsubscript{BERT}, on the other hand, fares worse than BERT\textsubscript{fine-tuned} and LORA, outperforming the former in only $19/48$ tasks and the latter in $25/48$ tasks, while exhibiting high standard deviations which can be explained by Theorem \ref{theorem_2}, as constraintSLP can behave erratically and not select the closest point to the class centroid. Overall, DeepSLP is the best performing method, demonstrating the highest accuracy in $31/48$ tasks, while being on-par with the second best model (BERT\textsubscript{fine-tuned}) on the remaining tasks ($15/48$ tasks). Fine-tuned BERT is, overall, the next best model with $13/48$ tasks while constraintSLP achieves the best performance amongst all methods in only $1/48$ tasks. ProtoNet's comparatively lower performance can be explained by the fact that meta-learning approaches tend to require a large number of diverse and structured meta-training tasks for effective learning --- thus not making them readily suited for (extreme) few-shot learning settings.

\paragraph{MetaSLP} In Table \ref{table:appendix_metaslp}, MetaSLP\textsubscript{REPTILE} outperforms all baselines achieving the highest performance in $33/48$ tasks. LEOPARD is the next best model with the highest performance in $11/48$ tasks. Interestingly, MetaSLP\textsubscript{FOMAML} does not fare as well as MetaSLP\textsubscript{REPTILE} and achieves the highest performance in only $1/48$ tasks while outperforms LEOPARD in only $6/48$ tasks. MetaSLP\textsubscript{FOMAML} nevertheless outperforms MetaSLP\textsubscript{REPTILE} and Reptile in natural language inference tasks -- demonstrating the usefulness of learnable inner-loop learning rates across multiple task distributions while meta-training.

{\small \begin{table*}
\centering
\scalebox{0.8}{
\def\arraystretch{1.1}
\begin{tabular}{lllllll}
\hline
\textbf{Category (Classes)} &\textbf{Shot} &\textbf{LORA\textsubscript{BERT}} & \textbf{ProtoNet} & \textbf{constraintSLP\textsubscript{BERT}} & \textbf{BERT\textsubscript{fine-tuned}*} & \textbf{DeepSLP\textsubscript{BERT}} \\
\hline
Political Bias (2) &4 &52.75 ± 4.33 &51.15 ± 2.454 &53.447 ± 3.281 &\colorbox{lightgray}{54.57 ± 5.02} &53.251 ± 4.042 \\
&8 &53.66 ± 4.25 &56.568 ± 4.228 &55.824 ± 3.725 &56.15 ± 3.75 &\colorbox{lightgray}{58.209 ± 5.198} \\
&16 &59.21 ± 2.27 &59.183 ± 4.706 &58.277 ± 4.128 &60.96 ± 4.25 &\colorbox{lightgray}{61.479 ± 2.974} \\
\hline
Emotion (13) &4 &7.56 ± 2.93 &8.953 ± 2.052 &8.662 ± 6.213 &09.20 ± 3.22 &\colorbox{lightgray}{9.076 ± 1.108} \\
&8 &9.02 ± 2.36 &\colorbox{lightgray}{10.857 ± 3.436} &8.16 ± 3.266 &08.21 ± 2.12 &8.041 ± 2.797 \\
&16 &10.29 ± 1.67 &11.479 ± 2.96 &8.115 ± 3.66 &\colorbox{lightgray}{13.43 ± 2.51} &10.919 ± 1.615 \\
\hline
Sentiment Books (2) &4 &51.27 ± 2.75 &55.53 ± 4.097 &59.89 ± 5.385 &54.81 ± 3.75 &\colorbox{lightgray}{58.67 ± 4.753} \\
&8 &58.16 ± 3.3 &58.97 ± 4.909 &64.34 ± 2.565 &53.54 ± 5.17 &\colorbox{lightgray}{64.78 ± 2.615} \\
&16 &59.16 ± 2.59 &65.5 ± 7.026 &66.36 ± 2.183 &65.56 ± 4.12 &\colorbox{lightgray}{67.453 ± 3.085} \\
\hline
Rating DVD (3) &4 &31.65 ± 4.91 &37.665 ± 7.184 &32.298 ± 16.263 &32.22 ± 08.72 &\colorbox{lightgray}{39.566 ± 5.086} \\
&8 &37.69 ± 3.16 &37.008 ± 5.118 &32.644 ± 16.016 &36.35 ± 12.50 &\colorbox{lightgray}{38.788 ± 4.449} \\
&16 &38.63 ± 5.52 &39.123 ± 6.004 &35.587 ± 17.445 &\colorbox{lightgray}{42.79 ± 10.18} &40.53 ± 4.375 \\
\hline
Rating Electronics (3) &4 &31.66 ± 2.94 &33.696 ± 5.55 &35.188 ± 16.211 &39.27 ± 10.15 &\colorbox{lightgray}{39.977 ± 5.959} \\
&8 &38.72 ± 5.95 &37.297 ± 5.938 &29.624 ± 12.876 &28.74 ± 08.22 &\colorbox{lightgray}{41.926 ± 3.985} \\
&16 &39.15 ± 6.6 &43.825 ± 5.946 &29.836 ± 12.753 &\colorbox{lightgray}{45.48 ± 06.13} &44.917 ± 3.164 \\
\hline
Rating Kitchen (3) &4 &36.63 ± 4.68 &35.914 ± 6.678 &28.253 ± 15.907 &34.76 ± 11.20 &\colorbox{lightgray}{39.624 ± 6.787} \\
&8 &39.69 ± 6.22 &38.46 ± 11.124 &24.397 ± 11.961 &34.49 ± 08.72 &\colorbox{lightgray}{41.081 ± 6.777} \\
&16 &38.17 ± 7.14 &46.546 ± 8.394 &31.926 ± 18.29 &\colorbox{lightgray}{47.94 ± 08.28} &45.801 ± 4.562 \\
\hline
Political Audience (2) &4 &49.75 ± 1.03 &50.976 ± 1.84 &51.305 ± 2.68 &51.02 ± 1.72 &\colorbox{lightgray}{51.741 ± 2.827} \\
&8 &54.05 ± 2.54 &52.022 ± 3.964 &53.104 ± 3.669 &52.80 ± 2.72 &\colorbox{lightgray}{54.506 ± 3.274} \\
&16 &55.39 ± 3.66 &54.024 ± 3.071 &53.888 ± 3.305 &\colorbox{lightgray}{58.45 ± 4.98} &56.956 ± 3.045 \\
\hline
Sentiment Kitchen (2) &4 &53.02 ± 1.54 &55.24 ± 3.427 &61.96 ± 4.594 &56.93 ± 7.10 &\colorbox{lightgray}{60.76 ± 4.426} \\
&8 &55.54 ± 3.47 &62.28 ± 5.103 &64.83 ± 3.983 &57.13 ± 6.60 &\colorbox{lightgray}{65.733 ± 3.198} \\
&16 &58.59 ± 4.83 &66.9 ± 5.441 &68.21 ± 3.298 &68.88 ± 3.39 &\colorbox{lightgray}{69.18 ± 2.589} \\
\hline
Disaster (2) &4 &56.02 ± 6.35 &51.474 ± 8.848 &52.77 ± 10.803 &\colorbox{lightgray}{55.73 ± 10.29} &54.252 ± 9.843 \\
&8 &57.46 ± 6.9 &60.661 ± 4.991 &56.888 ± 11.139 &56.31 ± 09.57 &\colorbox{lightgray}{61.3 ± 7.961} \\
&16 &65.79 ± 2.03 &63.893 ± 6.62 &65.907 ± 3.691 &64.52 ± 08.93 &\colorbox{lightgray}{69.28 ± 2.358} \\
\hline
Airline (3) &4 &24.36 ± 5.42 &44.167 ± 10.752 &36.243 ± 22.607 &42.76 ± 13.50 &\colorbox{lightgray}{50.987 ± 4.936} \\
&8 &52.31 ± 7.89 &50.148 ± 13.429 &44.972 ± 22.584 &38.00 ± 17.06 &\colorbox{lightgray}{55.209 ± 6.049} \\
&16 &54.1 ± 8.57 &54.8 ± 10.49 &29.238 ± 17.494 &58.01 ± 08.23 &\colorbox{lightgray}{60.247 ± 4.577} \\
\hline
Rating Books (3) &4 &34.69 ± 2.12 &37.715 ± 5.801 &25.562 ± 15.207 &39.42 ± 07.22 &\colorbox{lightgray}{42.116 ± 4.725} \\
&8 &39.36 ± 6.33 &38.518 ± 5.327 &34.026 ± 14.123 &39.55 ± 10.01 &\colorbox{lightgray}{42.156 ± 4.608} \\
&16 &41.23 ± 5.32 &44.694 ± 7.797 &32.509 ± 16.132 &43.08 ± 11.78 &\colorbox{lightgray}{46.513 ± 3.036} \\
\hline
Political Message (9) &4 &12.16 ± 1.46 &13.888 ± 2.076 &12.438 ± 1.799 &\colorbox{lightgray}{15.64 ± 2.73} &14.421 ± 1.095 \\
&8 &15.71 ± 2.04 &16.155 ± 2.316 &15.08 ± 2.925 &13.38 ± 1.74 &\colorbox{lightgray}{16.919 ± 1.756} \\
&16 &15.53 ± 2.55 &18.324 ± 2.011 &13.121 ± 3.294 &\colorbox{lightgray}{20.67 ± 3.89} &18.319 ± 1.74 \\
\hline
Sentiment DVD (2) &4 &50.77 ± 0.78 &51.06 ± 3.302 &\colorbox{lightgray}{56.06 ± 2.408} &54.98 ± 3.96 &55.003 ± 2.936 \\
&8 &52.24 ± 1.54 &55.19 ± 3.298 &56.98 ± 3.299 &55.63 ± 4.34 &\colorbox{lightgray}{57.527 ± 3.562} \\
&16 &52.6 ± 2.09 &59.45 ± 3.84 &58.95 ± 2.813 &58.69 ± 6.08 &\colorbox{lightgray}{60.76 ± 2.944} \\
\hline\hline
Scitail (2) &4 &43.36 ± 4.74 &50.227 ± 5.69 &52.296 ± 4.366 &\colorbox{lightgray}{58.53 ± 09.74} &54.101 ± 3.759 \\
&8 &54.29 ± 5.25 &54.196 ± 6.678 &55.964 ± 5.705 &\colorbox{lightgray}{57.93 ± 10.70} &56.341 ± 5.786 \\
&16 &52.68 ± 3.0 &57.744 ± 5.696 &59.675 ± 4.033 &\colorbox{lightgray}{65.66 ± 06.82} &59.692 ± 4.227 \\
\hline\hline
Restaurant (8) &4 &10.56 ± 1.36 &18.161 ± 2.822 &24.932 ± 17.102 &\colorbox{lightgray}{49.37 ± 4.28} &47.634 ± 5.237 \\
&8 &20.92 ± 2.4 &32.146 ± 5.785 &29.787 ± 9.573 &49.38 ± 7.76 &\colorbox{lightgray}{55.912 ± 4.494} \\
&16 &29.37 ± 4.05 &40.435 ± 3.348 &29.154 ± 13.537 &\colorbox{lightgray}{69.24 ± 3.68} &61.716 ± 2.208 \\
\hline
CoNLL (4) &4 &21.48 ± 2.71 &35.438 ± 7.324 &27.02 ± 7.346 &50.44 ± 08.57 &\colorbox{lightgray}{52.724 ± 5.84} \\
&8 &29.84 ± 3.28 &44.259 ± 4.886 &31.296 ± 17.487 &50.06 ± 11.30 &\colorbox{lightgray}{60.374 ± 3.731} \\
&16 &37.18 ± 3.32 &52.116 ± 5.354 &22.923 ± 7.933 &\colorbox{lightgray}{74.47 ± 03.10} &67.496 ± 4.551 \\
\hline
\end{tabular}}
\caption{
Classification performance (accuracy) of our methods (constraintSLP and DeepSLP) and baselines in the low-resource setting. Entries in grey indicate the best model out of all; * refers to the baseline as reported in \citet{leopard}. Subscripts for constraintSLP and DeepSLP refer to the (non-fine-tuned) encoder used. Each set of results is separated by a double line. The first set of results contains intent classification tasks, the second set has a natural language inference task and the last set contains entity typing tasks.}
\label{table:appendix_deepslp}
\end{table*}}

{\small \begin{table*}
\centering
\scalebox{0.85}{
\def\arraystretch{1.1}
\begin{tabular}{lllllll}
\hline
\textbf{Category (Classes)} &\textbf{Shot} &\textbf{ProtoNet} & \textbf{LEOPARD*} & \textbf{Reptile} & \textbf{MetaSLP\textsubscript{REPTILE}} & \textbf{MetaSLP\textsubscript{FOMAML}} \\
\hline
Political Bias (2) &4 &56.33 ± 4.37 &60.49 ± 6.66 &58.82 ± 4.31 &\colorbox{SpringGreen}{60.96 ± 6.13} &55.06 ± 5.9 \\
&8 &58.87 ± 3.79 &61.74 ± 6.73 &59.43 ± 3.79 &\colorbox{SpringGreen}{63.65 ± 4.57} &58.97 ± 5.5 \\
&16 &57.01 ± 4.44 &65.08 ± 2.14 &62.21 ± 0.72 &\colorbox{SpringGreen}{66.05 ± 1.57} &63.63 ± 4.74 \\
\hline
Emotion (13) &4 &09.18 ± 3.14 &11.71 ± 2.16 &11.65 ± 3.21 &\colorbox{SpringGreen}{11.94 ± 1.95} &11.03 ± 2.98 \\
&8 &11.18 ± 2.95 &12.90 ± 1.63 &10.56 ± 2.85 &\colorbox{SpringGreen}{13.42 ± 1.46} &12.38 ± 2.69 \\
&16 &12.32 ± 3.73 &13.38 ± 2.20 &11.62 ± 3.11 &\colorbox{SpringGreen}{14.03 ± 2.35} &12.32 ± 1.76 \\
\hline
Sentiment Books (2) &4 &73.15 ± 5.85 &82.54 ± 1.33 &76.95 ± 1.03 &\colorbox{SpringGreen}{83.22 ± 0.95} &74.51 ± 5.25 \\
&8 &75.46 ± 6.87 &83.03 ± 1.28 &77.49 ± 1.08 &\colorbox{SpringGreen}{83.8 ± 0.8} &79.25 ± 1.97 \\
&16 &77.26 ± 3.27 &83.33 ± 0.79 &77.88 ± 0.56 &\colorbox{SpringGreen}{83.8 ± 1.59} &78.41 ± 1.08 \\
\hline
Rating DVD (3) &4 &47.73 ± 6.20 &\colorbox{SpringGreen}{49.76 ± 9.80} &45.91 ± 9.85 &45.2 ± 8.91 &39.64 ± 5.17 \\
&8 &47.11 ± 4.00 &53.28 ± 4.66 &47.23 ± 9.22 &\colorbox{SpringGreen}{58.38 ± 2.9} &52.35 ± 5.27 \\
&16 &48.39 ± 3.74 &53.52 ± 4.77 &48.49 ± 8.88 & 57.41 ± 4.71&\colorbox{SpringGreen}{60.4 ± 3.71} \\
\hline
Rating Electronics (3) &4 &37.40 ± 3.72 &\colorbox{SpringGreen}{51.71 ± 7.20} &44.47 ± 8.25 &45.34 ± 7.22 &39.53 ± 5.76 \\
&8 &43.64 ± 7.31 &54.78 ± 6.48 &49.1 ± 6.81 &\colorbox{SpringGreen}{55.10 ± 5.12} &47.83 ± 5.94 \\
&16 &44.83 ± 5.96 &58.69 ± 2.41 &50.68 ± 6.8 &\colorbox{SpringGreen}{59.47 ± 2.29} &56.53 ± 4.36 \\
\hline
Rating Kitchen (3) &4 &44.72 ± 9.13 &\colorbox{SpringGreen}{50.21 ± 09.63} &45.38 ± 10.96 &45.20 ± 8.78 &39.11 ± 7.16 \\
&8 &46.03 ± 8.57 &53.72 ± 10.31 &46.71 ± 9.84 &\colorbox{SpringGreen}{54.53 ± 9.9} &50.19 ± 8.36 \\
&16 &49.85 ± 9.31 &57.00 ± 08.69 &52.87 ± 9.52 &\colorbox{SpringGreen}{58.94 ± 7.58} &57.63 ± 8.37 \\
\hline
Political Audience (2) &4 &51.47 ± 3.68 &52.60 ± 3.51 &52.45 ± 4.26 &\colorbox{SpringGreen}{54.1 ± 3.66} &52.03 ± 2.73 \\
&8 &51.83 ± 3.77 &54.31 ± 3.95 &52.87 ± 4.31 &\colorbox{SpringGreen}{56.01 ± 3.65} &52.06 ± 2.27 \\
&16 &53.53 ± 3.25 &57.71 ± 3.52 &55.6 ± 1.85 &\colorbox{SpringGreen}{58.57 ± 2.04} &54.33 ± 3.14 \\
\hline
Sentiment Kitchen (2) &4 &62.71 ± 9.53 &78.35 ± 18.36 &69.81 ± 14.58 &\colorbox{SpringGreen}{81.96 ± 3.73} &72.73 ± 7.97 \\
&8 &70.19 ± 6.42 &\colorbox{SpringGreen}{84.88 ± 1.12} &75.76 ± 1.13 &83.33 ± 1.99 &76.86 ± 4.46 \\
&16 &71.83 ± 5.94 &\colorbox{SpringGreen}{85.27 ± 1.31} &76.41 ± 0.66 &84.33 ± 1.81 &80.78 ± 4.38 \\
\hline
Disaster (2) &4 &50.87 ± 1.12 &51.45 ± 4.25 &49.76 ± 4.73 &\colorbox{SpringGreen}{55.03 ± 8.73} &52.62 ± 2.71 \\
&8 &51.30 ± 2.30 &55.96 ± 3.58 &52.17 ± 5.17 &\colorbox{SpringGreen}{57.77 ± 6.40} &55.04 ± 5.79 \\
&16 &52.76 ± 2.92 &61.32 ± 2.83 &55.37 ± 4.53 &\colorbox{SpringGreen}{65.18 ± 4.41} &62.27 ± 4.42 \\
\hline
Airline (3) &4 &40.27 ± 8.19 &54.95 ± 11.81 &57.11 ± 14.16 &\colorbox{SpringGreen}{57.39 ± 7.83} &51.62 ± 10.53 \\
&8 &51.16 ± 7.60 &61.44 ± 03.90 &64.37 ± 3.49 &\colorbox{SpringGreen}{65.67 ± 4.82} &57.47 ± 9.37 \\
&16 &48.73 ± 6.79 &62.15 ± 05.56 &66.31 ± 2.55 &\colorbox{SpringGreen}{69.48 ± 2.06} &65.02 ± 5.16 \\
\hline
Rating Books (3) &4 &48.44 ± 7.43 &54.92 ± 6.18 &\colorbox{SpringGreen}{56.57 ± 8.17} &55.79 ± 5.61 &54.4 ± 5.83 \\
&8 &52.13 ± 4.79 &59.16 ± 4.13 &57.33 ± 7.63 &\colorbox{SpringGreen}{65.74 ± 5.58} &57.17 ± 6.77 \\
&16 &57.28 ± 4.57 &61.02 ± 4.19 &63.26 ± 3.59 &\colorbox{SpringGreen}{67.87 ± 3.45} &66.66 ± 3.93 \\
\hline
Political Message (9) &4 &14.22 ± 1.25 &15.69 ± 1.57 &14.58 ± 1.78 &\colorbox{SpringGreen}{18.84 ± 1.82} &14.96 ± 1.94 \\
&8 &15.67 ± 1.96 &18.02 ± 2.32 &15.13 ± 2.16 &\colorbox{SpringGreen}{20.09 ± 2.71} &16.09 ± 2.6 \\
&16 &16.49 ± 1.96 &18.07 ± 2.41 &16.38 ± 2.15 &\colorbox{SpringGreen}{23.22 ± 1.17} &16.62 ± 2.19 \\
\hline
Sentiment DVD (2) &4 &74.38 ± 2.44 &80.32 ± 1.02 &72.03 ± 11.61 &\colorbox{SpringGreen}{80.97 ± 1.21} &73.08 ± 7.56 \\
&8 &75.19 ± 2.56 &80.85 ± 1.23 &75.79 ± 1.62 &\colorbox{SpringGreen}{81.85 ± 1.79} &76.55 ± 2.9 \\
&16 &75.26 ± 1.07 &81.25 ± 1.41 &76.69 ± 0.8 &\colorbox{SpringGreen}{83.48 ± 1.01} &78.19 ± 1.32 \\
\hline\hline
Scitail (2) &4 &\colorbox{SpringGreen}{76.27 ± 4.26} &69.50 ± 9.56 &59.13 ± 10.58 &53.48 ± 5.59 &61.55 ± 9.11 \\
&8 &\colorbox{SpringGreen}{78.27 ± 0.98} &75.00 ± 2.42 &62.63 ± 10.85 &60.79 ± 4.6 &68.03 ± 4.54 \\
&16 &\colorbox{SpringGreen}{78.59 ± 0.48} &77.03 ± 1.82 &68.03 ± 1.57 &61.67 ± 3.61 &68.5 ± 3.7 \\
\hline\hline
Restaurant (8) &4 &17.36 ± 2.75 &\colorbox{SpringGreen}{49.84 ± 3.31} &13.37 ± 2.25 &27.00 ± 2.61 &20.31 ± 2.97 \\
&8 &18.70 ± 2.38 &\colorbox{SpringGreen}{62.99 ± 3.28} &16.83 ± 3.42 &35.66 ± 2.39 &27.74 ± 2.29 \\
&16 &16.41 ± 1.87 &\colorbox{SpringGreen}{70.44 ± 2.89} &16.0 ± 3.44 &37.20 ± 2.68 &28.57 ± 2.41 \\
\hline
CoNLL (4) &4 &32.23 ± 5.10 &\colorbox{SpringGreen}{54.16 ± 6.32} &31.31 ± 5.32 &40.79 ± 3.40 &36.07 ± 3.25 \\
&8 &34.49 ± 5.15 &\colorbox{SpringGreen}{67.38 ± 4.33} &33.17 ± 5.1 &41.25 ± 5.21 &40.5 ± 2.16 \\
&16 &33.75 ± 6.05 &\colorbox{SpringGreen}{76.37 ± 3.08} &34.04 ± 3.59 &45.96 ± 4.75 &43.67 ± 6.92 \\
\hline
\end{tabular}}
\caption{
Classification performance (accuracy) of MetaSLP and baselines in the high-resource setting. Entries in green indicate the best model out of all; * refers to the baseline as reported in \citet{leopard}. Each set of results is separated by a double line. The first set of results contains intent classification tasks, the second set has a natural language inference task and the last set contains entity typing tasks.}
\label{table:appendix_metaslp}
\end{table*}}

\subsection{Ensemble properties of DeepSLP}
\label{appendix:ensemble}

In this study, we compare and contrast DeepSLP to ensembles and draw similarities between the two, shedding further light into the effectiveness of our approach. 
Each prediction decision by DeepSLP is the result of two soft-label prototypes -- those that lie on each end of the line nearest to a test point $x$.  
An analogy can then be drawn between the prototypes used at prediction time and those individual (albeit independent) models that are utilised by an ensemble when producing the final classification.  

While DeepSLP prototypes are not independent but are rather trained jointly (and share the same encoder), in what follows, we demonstrate that they display several properties of ensemble methods, while being computationally efficient and utilising a small number of parameters. 
For the analyses below, we consider the tasks Airline and Disaster using an 8-shot setting and evaluate on the test data for each. However, we find the below properties to generalise across all tasks. 

\subsubsection{Individual vs joint prediction}

In a similar way as an ensemble exhibits superior performance to the individual models it utilises, we seek to assess whether the joint utilisation of prototypes at prediction time is indeed more effective than utilising each prototype individually.  
To evaluate this, we measure the probability distribution of each setting on the test data using negative log-likelihood:

$$
NLL(\mathbf{f(x)},y) \triangleq -log(\mathbf{f^{(y)}(x)})    
$$
Following \citet{ensembles_work}, for a strictly convex function such as $NLL$, we use Jensen's inequality:  
$$
NLL(\mathbf{F(x)},y) \; \leq \; \mathbb{E}[NLL(\mathbf{f(x)}, y)]   
$$
\begin{figure}[t]
    \centering
     \begin{subfigure}{\columnwidth}
         \includegraphics[width=\columnwidth]{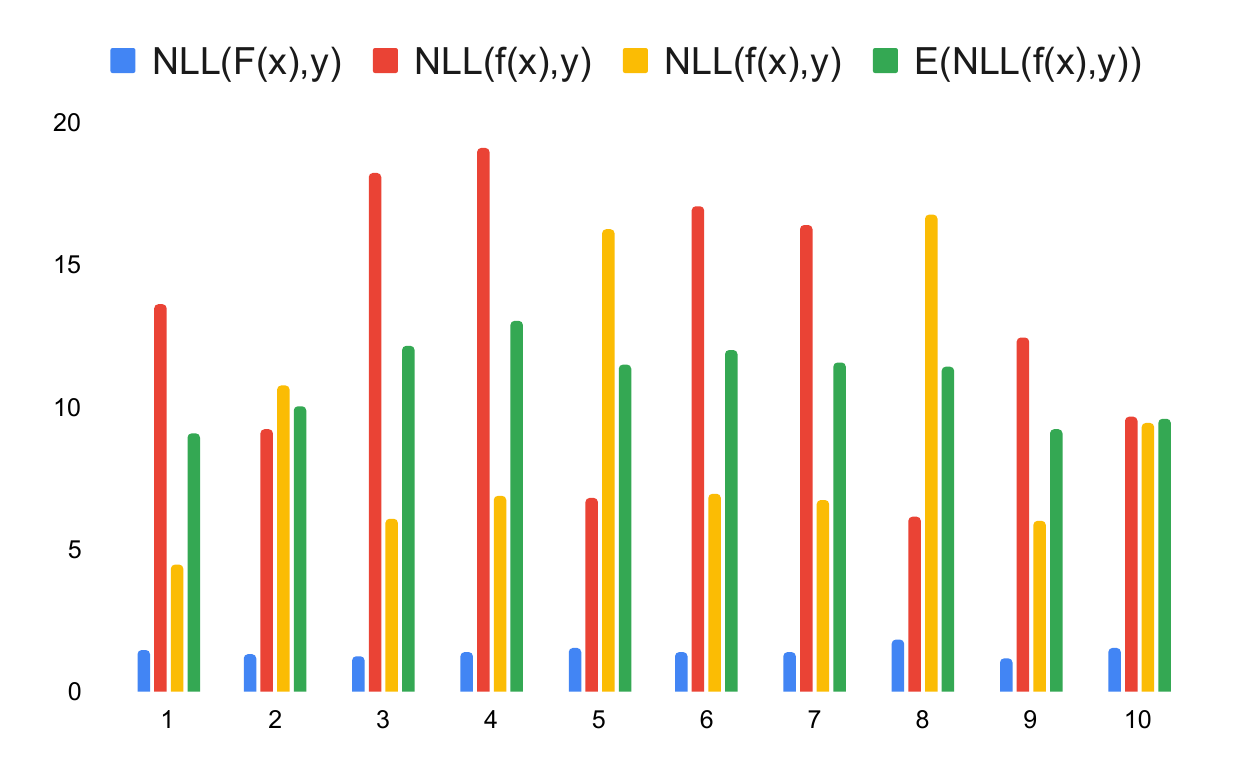}
     \end{subfigure}
     \hfill
     \begin{subfigure}{\columnwidth}
         \includegraphics[width=\columnwidth]{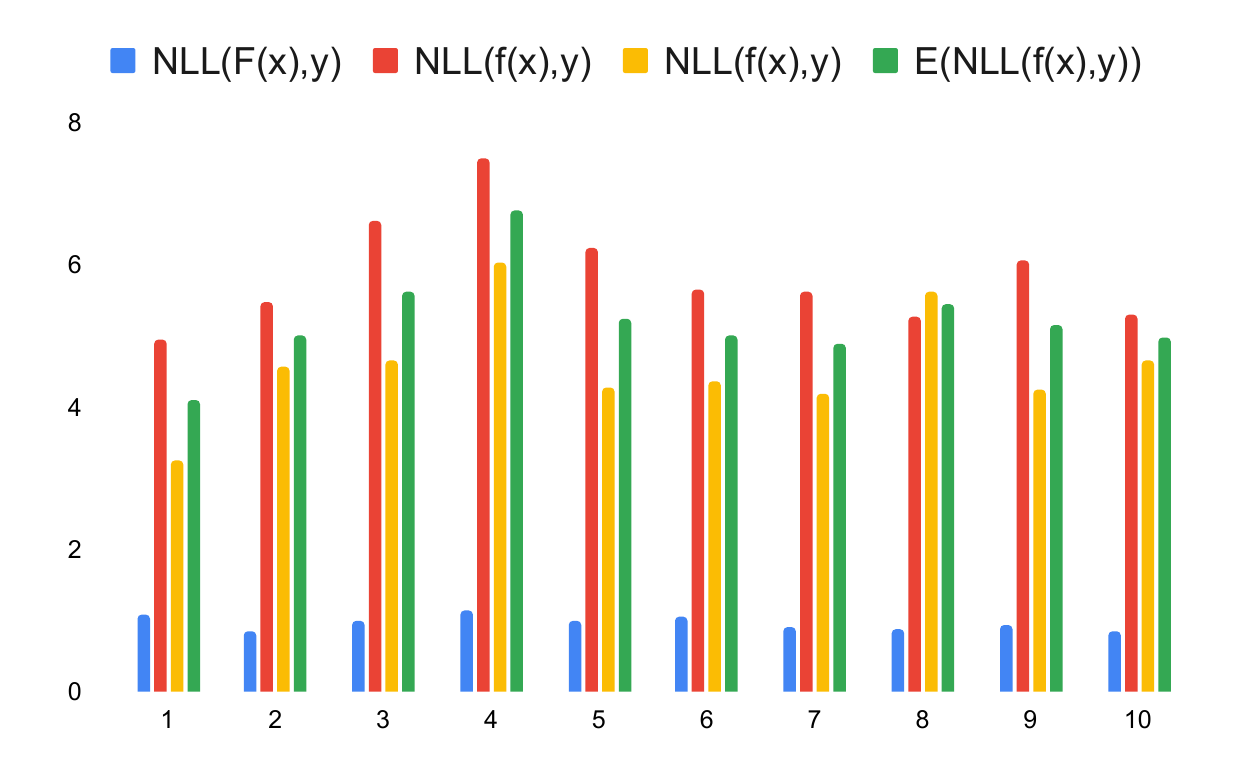}
     \end{subfigure}
\caption{$NLL(\mathbf{F(x)},y)$ vs. $\mathbb{E}[NLL(\mathbf{f(x)}, y)]$ for {Disaster} (top) and {Airline} (bottom). The \textit{i\textsuperscript{th}} subscript refers to the \textit{i\textsuperscript{th}} soft-label prototype.}
\label{fig:property_1}
\end{figure}
where $F(x)$ is the ensemble and $f(x)$ are the constituent models. The idea is that the probability distribution of the ensemble fits the target distribution more closely than the corresponding expected probability distributions of its constituent models. For joint soft-label prototypes parameterised by $g_1$ and $g_2$ and located at $p_1$ and $p_2$, we have: 
{\footnotesize $$
\begin{aligned}
NLL\left(softmax\left(\frac{g_1(f(x))}{|| f(x) - p_1 ||}) + \frac{g_2(f(x))}{|| f(x) - p_2 ||}\right), y\right) \;\;\;\;\;\\ 
= - \sum log\left(softmax\left(\frac{g_1^y(f(x))}{|| f(x) - p_1 ||}) + \frac{g_2^y(f(x))}{|| f(x) - p_2 ||}\right)\right) \;
\end{aligned} 
$$}

For the individual soft-label prototypes, weighing the outputs by distance does not change the final softmax probability distribution; therefore, we can define $\mathbb{E}[NLL(\mathbf{f(x)}, y)]$ as their average:  
{\footnotesize $$
\begin{aligned}
-\frac{1}{2}\sum log(softmax(g_1^y(f(x)) + log(softmax(g_2^y(f(x))    
\end{aligned}
$$}

\noindent We plot the negative log likelihoods for Airline and Disaster on the test set after fine-tuning each model on ten subsets of few-shot training data (as explained before in Section \ref{datasets}) in Figure \ref{fig:property_1} to assess whether DeepSLP exhibits this property of ensemble methods. We find that $NLL(\mathbf{F(x)},y)$ is much lower than $\mathbb{E}[NLL(\mathbf{f(x)}, y)]$, which confirms that the joint utilisation of prototypes results in better predictions than if they were to be used individually. Our experiments confirm that this is a general trend we observe across tasks. In the following sections, we investigate the reasons behind the high values observed for $\mathbb{E}[NLL(\mathbf{f(x)}, y)]$, and compare the jointly trained prototypes against a strong baseline, the fine-tuned BERT baseline. 

\subsubsection{Jointly utilised soft-label prototypes improve diversity}

Diversity in ensemble classifications refers to the difference in the probability distribution on out-of-distribution (ood) data for classifications between individual models and the ensemble. We use this definition to ascertain the diversity of classifications provided by the jointly utilised soft-label prototypes. Diversity is a desirable property as ensemble predictions are generally more robust due to diversity between the predictions of their individual members \cite{why_m_heads}.

Existing work \cite{ashukha2020pitfalls, balaji} defines ensemble uncertainty as the sum of ensemble diversity and the expected average model uncertainty on ood data. Based on \citet{ensembles_work}, it is calculated as:
$$
    H([y|F(x)] = \frac{-1}{C}\sum p(y_i | F(x))log(p(y_i | F(x)))
$$
If we use the Jenson-Shannon divergence as a diversity measure for an ensemble given by $$JSD_{p(f)} [y |f(x)] = \frac{1}{M}\sum KL[y | f(x) || y | F(x)]$$ where KL is the average KL divergence between the output distribution of each soft-label prototype and the jointly utilised soft-label prototypes, from \citet{ensembles_work}, this expression reduces to:
$ {\textstyle
    H([y|F(x)] = \stackrel{\underline{ens.\;diversity}}{JSD_{p(f)}[y |f(x)]} +  \stackrel{\underline{avg.\;model\;uncert.}}{E_{p(f)}[H [y |f(x)]]}
} $

\begin{figure}[t]
 \centering
 \includegraphics[width=0.5\textwidth]{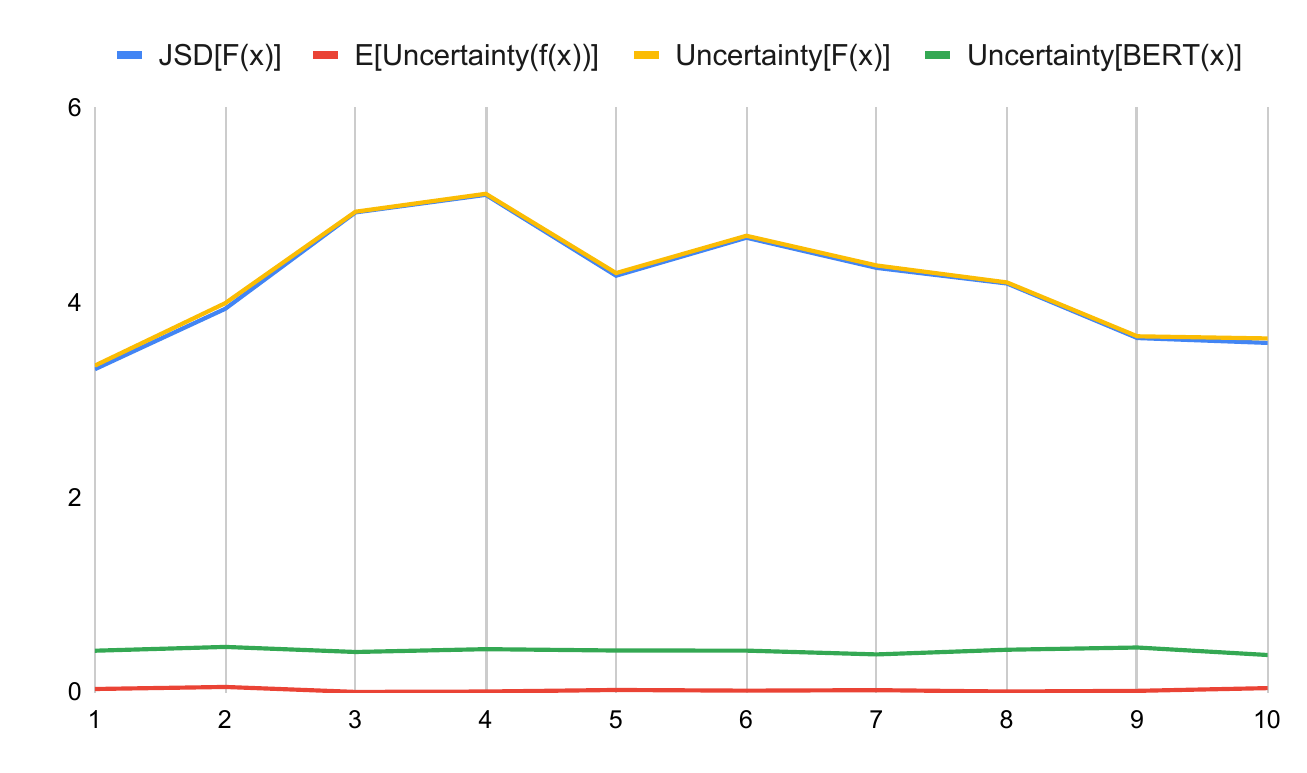}
 \caption{Ensemble uncertainty contrasted against the uncertainty of fine-tuned BERT, where we observe that DeepSLP's uncertainty $F(x)$ (given by yellow) is driven by ensemble diversity, given by $JSD(F(x))$ in blue.}
\label{fig:property_3}
\end{figure}

\paragraph{}We contrast ensemble uncertainty and single model uncertainty using the fine-tuned BERT model for the task \textit{airline} in Figure \ref{fig:property_3}, but note that similar trends are observed across all tasks. 
We note that the uncertainty of jointly utilised soft-label prototypes is generally higher than that of the fine-tuned BERT model. As the average model uncertainty of individual soft-label prototypes is negligibly low, the uncertainty in the joint case is driven mainly by the diversity of the ensemble. This is in line with previous work which attributes an increase in uncertainty in ensembles due to diversity \citep{balaji, ensemble_ml, wilson}. Though not strictly an ensemble, our approach exhibits similar properties (higher uncertainty driven by model diversity), as well as a general reduction in standard deviation compared to fine-tuned BERT. 

The above provide evidence that our approach as a whole exhibits desirable properties of ensembles which drive a higher performance but which do not lead to higher training time nor compute. 

\end{document}